\documentclass[10pt]{article} 
\usepackage[accepted]{tmlr}

\usepackage{amsmath,amsfonts,bm}

\def\eqref#1{equation~\ref{#1}}

\def\1{\bm{1}}

\DeclareMathAlphabet{\mathsfit}{\encodingdefault}{\sfdefault}{m}{sl}
\SetMathAlphabet{\mathsfit}{bold}{\encodingdefault}{\sfdefault}{bx}{n}

\usepackage{hyperref}       
\usepackage{url}            
\usepackage{booktabs}       
\usepackage{amsmath}        
\usepackage{amsfonts}       
\usepackage{nicefrac}       
\usepackage{microtype}      
\usepackage{xcolor}         
\usepackage{graphicx}      %

\usepackage{wrapfig}

\title{Structure Over Scale: Rethinking Adaptation \\
for Reinforcement Learning with Verifiable Rewards}

\author{\name Allan Kazakov \email allan.kazakov@bahcesehir.edu.tr \\
      \addr Department of Artificial Intelligence\\
      Bahçeşehir University
      \AND
      \name Abdurrahman Javat \email abdurrahman.javat@bahcesehir.edu.tr \\
      \addr Department of Artificial Intelligence\\
      Bahçeşehir University
      }

\def\month{10}  
\def\year{2026} 
\def\openreview{\url{https://openreview.net/forum?id=17kr8cCW7s}} 

\begin{document}

\maketitle

\begin{abstract}
Reinforcement Learning with Verifiable Rewards (RLVR) is often framed as a
choice between the expressivity of Full Fine-Tuning (FFT) and the efficiency of
Low-Rank Adaptation (LoRA). We find that this framing misses an important
variable: the geometry of the update. Under a common cold-start GRPO protocol, LoRA exhibits \emph{rank collapse}:
performance remains broadly stable at low rank, then fails abruptly as rank
increases. On DeepMath-Hard, for example, accuracy drops from $73.1\%$ at
$r=64$ to $4.7\%$ at $r=128$. In contrast, Weight-Decomposed Low-Rank
Adaptation (DoRA) and Quantum-Informed Tensor Adaptation (QuanTA) remain stable
over a wider tested optimization region and are competitive with FFT while
updating less than $0.6\%$ of the model parameters. Spectral analysis reveals a corresponding difference in learned update
structure: collapsed LoRA concentrates its update energy into relatively few
effective directions, whereas structured adapters produce substantially more
distributed updates. These results suggest that nominal rank and parameter
count alone are insufficient to characterize adaptation in RLVR --- the structure
imposed on the reachable update manifold also matters. Finally, across three 8B base models, adapter behavior aligns with both
frozen-weight spectral structure and reported pre-training scale, motivating
the \emph{Model Maturity Hypothesis}: the adaptation geometry that works best
may depend on properties already encoded during pre-training. We present this
as a falsifiable hypothesis rather than a causal conclusion. The code can be found here \href{https://github.com/Shrimpstanot/structure-over-scale}{\texttt{https://github.com/Shrimpstanot/structure-over-scale}}.
\end{abstract}

\section{Introduction}
\label{sec:intro}
 
\subsection{The RLVR Efficiency Problem}
 
The dominant recipe for training reasoning-capable large language models
(LLMs) pairs a base model pre-trained on trillions of tokens with a
Reinforcement Learning with Verifiable Rewards (RLVR) post-training
stage~\citep{deepseek2025r1}.
Within this stage, Group Relative Policy Optimization
(GRPO)~\citep{Shao2024DeepSeekMath} has emerged as the standard method for 
reasoning: it eliminates a separate value-function critic by normalizing
rewards within a group of sampled rollouts, halving the memory footprint
compared with PPO.
 
Despite GRPO's efficiency, the field faces a persistent dichotomy when
choosing how many model parameters to update.
Leading laboratories default to \emph{Full Fine-Tuning} (FFT)~\citep{deepseek2025r1}, accepting
massive memory and gradient-synchronization overhead under the assumption
that reasoning updates require high intrinsic dimensionality.
Academic work instead adopts \emph{Low-Rank Adaptation}
(LoRA)~\citep{Hu2021LoRA}, citing evidence that rank $r{=}16$ achieves
performance parity with FFT when hyperparameters are carefully
optimized~\citep{schulman2025lora}.
The field has thus settled into a binary choice: expensive correctness or
affordable approximation.
 
We present experimental evidence that this binary is incomplete, and study structured high-rank adaptation as a third regime.
 
\subsection{Two Surprising Findings}
 
\paragraph{Finding 1 — Rank Collapse in RLVR.}
The SFT literature predicts that increasing LoRA rank yields diminishing
returns \citep{Hu2021LoRA,albert2025randlora}.
Our LoRA rank sweep on Qwen~3 8B trained with GRPO on DeepMath-Hard
reveals a qualitatively different phenomenon:
 
\begin{table}[h]
  \centering
  \caption{LoRA rank sweep on Qwen~3 8B under the common cold-start GRPO
  protocol (DeepMath-Hard, seed 42, batch size 64, generations 8). Performance remains
  broadly stable throughout the low-rank regime before an abrupt collapse
  between $r=64$ and $r=128$.}
  \label{tab:intro_rank_sweep}
  \small
  \begin{tabular}{lcccccccc}
    \toprule
    Nominal rank $r$ & 1 & 4 & 8 & 16 & 32 & 64 & 128 & 256 \\
    \midrule
    Accuracy & 79.29\% & 79.29\% & 78.10\% & 76.20\% &
               77.30\% & 73.10\% & \textbf{4.70\%} & \textbf{2.30\%} \\
    \bottomrule
  \end{tabular}
\end{table}
 
\noindent Performance is broadly stable across low ranks, declines at
$r=64$, and then collapses abruptly at $r\geq128$ under the common protocol.
The failure is not removed by simply increasing global batch size: a
second-seed replication reproduces the key rank--batch boundary, with
$r=64$ remaining functional at batch size 64 but collapsing at batch sizes
128 and 256 in both seeds, while $r=256$ remains near floor throughout.
Varying the GRPO group size over $G\in\{4,8,16\}$ can affect performance,
but likewise does not rescue $r=256$ from the collapsed regime.
In contrast, DoRA and QuanTA remain trainable over a wider
tested optimization region.

SVD analysis (Figure~\ref{fig:effective_rank}) identifies a corresponding spectral
signature: collapsed high-rank LoRA uses only a small fraction of its nominal
rank, with update energy concentrated into relatively few effective
directions, whereas the structured adapters produce substantially more
distributed updates. Effective rank alone does not determine policy quality,
but this divergence shows that increasing nominal rank does not necessarily
translate into increased usable update dimensionality under RLVR optimization.
This behavior contrasts with the diminishing-returns pattern reported in the
SFT literature we compare against.
 
\paragraph{Finding 2 — The Model Maturity Hypothesis.}
Alongside rank collapse, we observe a second partition across base models.
Among the model-level quantities measured in our study, reported pre-training
scale aligns with both the observed adapter behavior and the spectral structure
of the frozen base weights.
Table~\ref{tab:maturity_teaser} summarizes the behavioral pattern across three
architecturally independent 8B models.
 
\begin{table}[h]
  \centering
  \caption{Performance split across three base models.
           All results on DeepMath-Hard (Seed 42).
           The two ${\sim}15$T-token models exhibit similar adapter behavior,
while Qwen~3~8B shows a substantially larger structured-adapter gap.}
  \label{tab:maturity_teaser}
  \small
  \begin{tabular}{lccccc}
    \toprule
    Model        & Pre-training & Best LoRA & Best QuanTA & $\Delta$ \\
    \midrule
    Apertus 8B   & ${\sim}15$T  & 2.3\%     & 3.5\%          & $+$1.2\% \\
    Llama 3.1 8B & ${\sim}15.6$T & 2.7\%    & 3.9\%          & $+$1.2\% \\
    Qwen 3 8B & ${\sim}36$T & 79.29\% & 84.0\% & $+$4.71\% \\
    \bottomrule
  \end{tabular}
\end{table}
 
\noindent The two ${\sim}15$T models not only agree in direction but also
exhibit a qualitative signature absent from Qwen~3. 
We hypothesize that this behavioral bifurcation is related to pre-training
scale. The 36T-token model has markedly flatter singular value spectra in its
frozen weights, suggesting more distributed pre-trained representations. Under
this interpretation, adaptation may benefit from updates spread across many
singular directions. We present this as a falsifiable empirical hypothesis rather
than an established causal law: the two groups are cleanly separated, but
architecture, tokenizer, and data mixture co-vary with pre-training scale and
cannot be fully disentangled with the current model set.
 
\subsection{Why Structure, Not Rank Alone}
 
Both findings point to the same organizing principle. Rank collapse shows that
increasing nominal update capacity is not monotonically beneficial in
cold-start RLVR under our fixed-pass GRPO protocol. The model-maturity results
further suggest that the useful geometry of an update may depend on structure
already present in the pre-trained model. The resolution we study is therefore
not simply low rank versus high rank, but structured adaptation: retaining
distributed update capacity while restricting which solutions gradient descent
can reach.

\subsection{Contributions}
\begin{enumerate}

  \item \textbf{Rank Collapse as an Optimization-Stability Boundary.}
  We systematically sweep LoRA rank over
  $r \in \{1,4,8,16,32,64,128,256\}$ and identify an abrupt transition from
  a broadly stable low-rank regime to near-floor performance at high rank
  under a common cold-start GRPO protocol. The key rank--batch boundary
  reproduces across two seeds, while increasing either global batch size or
  GRPO group size within the tested ranges does not rescue collapsed
  high-rank LoRA.

  \item \textbf{The Full Fine-Tuning Effective-Rank Paradox.}
  We show that FFT can produce substantially lower effective-rank learned
  updates than structured adapters using less than 0.6\% of the model
  parameters. Collapsed LoRA likewise uses only a small fraction of its
  nominal rank, demonstrating that trainable parameter count and nominal rank
  alone do not characterize the geometry actually reached by optimization.

  \item \textbf{Controlled Comparison with Spectral Diagnostics.}
  We provide a head-to-head empirical comparison of LoRA, DoRA, QuanTA, and
  FFT under a shared cold-start GRPO pipeline, including multi-seed
  evaluation, rank and batch-size sweeps, a GRPO group-size ablation, and
  SVD-based diagnostics across the resulting checkpoints. Structured adapters
  remain competitive with FFT on the primary multi-seed comparison while
  updating orders of magnitude fewer parameters and occupying a wider tested
  stability region than high-rank LoRA.

  \item \textbf{The Model Maturity Hypothesis.}
  We identify a concordant behavioral and frozen-weight spectral partition
  across three independently developed 8B models spanning approximately 15T
  to 36T reported pre-training tokens. Reported pre-training scale aligns with
  this partition in the current model set, motivating the falsifiable
  hypothesis that the update geometry most useful for RLVR may depend on
  structure acquired during pre-training.

  \item \textbf{Practical Guidance for Efficient RLVR.}
  Our results reframe the FFT-versus-LoRA choice around update geometry rather
  than parameter count alone. Under the tested cold-start GRPO protocol,
  unconstrained high-rank LoRA can fail abruptly; larger batches should not be
  assumed to restore stability, and larger GRPO groups can help some
  configurations without rescuing the collapsed high-rank regime. Structured
  adapters provide a parameter-efficient alternative that remains functional
  over a wider tested optimization region.

\end{enumerate}


\section{Background and Related Work}
\label{sec:background}

\subsection{Reinforcement Learning with Verifiable Rewards}
\label{sec:background:rlvr}

RLVR uses ground-truth outcomes—such as mathematically verified answers—as the
reward signal, avoiding the over-optimization risks of learned reward models
\citep{deepseek2025r1}.
\citet{Shao2024DeepSeekMath} introduced Group Relative Policy Optimization
(GRPO), which eliminates the memory cost of a separate critic by computing
per-output advantages within a group of $G$ completions sampled from the current
policy:
\begin{equation}
  A_i = \frac{r_i - \operatorname{mean}(r_1, \ldots, r_G)}
             {\operatorname{std}(r_1, \ldots, r_G) + \epsilon},
  \label{eq:grpo_advantage}
\end{equation}
where $r_i \in \{0, 1\}$ is the binary correctness reward for the $i$-th
completion.
Sparse binary rewards can yield noisy or uninformative within-group optimization
signals; Section~\ref{sec:experiments:batch_size} tests whether sampling scale
alone can account for the observed collapse.
We operate in the cold-start (Base-to-RL) regime \citep{liu2025understanding},
in which the model must simultaneously learn output structure and reasoning logic
from the reward signal alone, without any SFT warmup.
We adopt the Dr.~GRPO token-level loss normalization \citep{liu2025understanding}
to prevent penalizing longer Chain-of-Thought completions.

\subsection{Parameter-Efficient Fine-Tuning}
\label{sec:background:peft}

\paragraph{LoRA.}
\citet{Hu2021LoRA} freezes the pre-trained weight matrix $W_0 \in \mathbb{R}^{d
\times k}$ and injects a trainable low-rank update $\Delta W = BA$, where $B \in
\mathbb{R}^{d \times r}$, $A \in \mathbb{R}^{r \times k}$, and $r \ll
\min(d,k)$:
\begin{equation}
  W = W_0 + \Delta W, \qquad \Delta W = BA.
  \label{eq:lora}
\end{equation}
The method is grounded in the Intrinsic Dimension Hypothesis
\citep{aghajanyan2020intrinsic}, which posits that downstream adaptation lies in
a low-dimensional subspace—a characterization established in SFT settings whose
validity in RLVR our experiments directly challenge.


\paragraph{DoRA.}
\citet{liu2024dora} decompose each weight matrix into a magnitude scalar and a
directional component, applying a LoRA update only to the direction:
\begin{equation}
  W = \lVert W_0 \rVert_c \cdot
      \frac{W_0 + \Delta W}{\lVert W_0 + \Delta W \rVert_c},
  \label{eq:dora}
\end{equation}
where $\lVert \cdot \rVert_c$ denotes column-wise norm.
By forcing the optimizer to treat magnitude and direction as independent degrees
of freedom, this decomposition acts as a structural constraint on the update
manifold—one that, as our spectral analysis shows, produces substantially higher
effective rank than LoRA at identical nominal rank.
DoRA is therefore not simply a LoRA variant with better hyperparameters: it is
a structurally constrained adapter whose inductive bias places it in the same
mechanistic category as QuanTA, achieved via a different parameterization.

\paragraph{QuanTA.}
\citet{chen2024quanta} parameterize $\Delta W$ as a Matrix Product Operator
(MPO), a tensor network structure adapted from quantum many-body physics.
For a weight matrix with reshaped local indices $\mathbf{i} = (i_1, \ldots,
i_n)$ and $\mathbf{j} = (j_1, \ldots, j_n)$, the update takes the form:
\begin{equation}
  \Delta W(\mathbf{i}, \mathbf{j}) =
    \sum_{\boldsymbol{\alpha}}
    \prod_{k=1}^{n} C^{(k)}_{\alpha_{k-1} \alpha_k}(i_k, j_k),
  \label{eq:quanta}
\end{equation}
where $C^{(k)} \in \mathbb{R}^{\chi \times \chi \times d_k \times d_k}$ are
learnable core tensors and $\chi$ is the bond dimension.
The MPO structure decouples adaptation rank from parameter count: $\Delta W$ can
approximate a full-rank transformation while parameters scale linearly in the
number of modes $n$.
We evaluate two configurations: a parameter efficient setting ($d=4$, decomposition
$[16,8,8,4]$, 7.7M parameters) and a high-capacity setting ($d=3$,
decomposition $[16,16,16]$, 35.3M parameters).
BOFT \citep{liu2023boft} and MoRA \citep{jiang2024mora} impose alternative
structural constraints; we defer a conceptual geometric comparison to
Appendix~\ref{app:peft_theory}.

\subsection{Related Work}
\label{sec:background:related}

\paragraph{Rank and expressivity in SFT.}
The SFT literature consistently finds that increasing LoRA rank yields
diminishing returns, with performance plateauing well below the full-rank
regime \citep{Hu2021LoRA,liu2024dora}—a pattern explained by the Intrinsic
Dimension Hypothesis \citep{aghajanyan2020intrinsic}.
Our findings establish a qualitative distinction under the studied cold-start
GRPO protocol: sufficiently high LoRA rank can produce an abrupt performance
collapse rather than merely diminishing returns.

\paragraph{Stability and scheduling in RLVR.}
\citet{schulman2025lora} report that Base-to-RL fine-tuning with LoRA $r=16$
achieves parity with FFT under optimized hyperparameters, that constant
learning-rate schedules are stable at large batch sizes, and that high-rank LoRA
does not collapse in their setup.
Our results show that these conclusions do not transfer unchanged to our
fixed-pass cold-start GRPO protocol: constant schedules collapse at batch size
64, LoRA parity depends on the structured-adapter baseline, and increasing batch
size alone does not rescue high-rank LoRA under our fixed protocol.
We therefore attribute the discrepancy to the joint effects of batch size,
update horizon, learning-rate schedule, and optimization protocol rather than to
a single universal LoRA rank threshold.

\paragraph{PEFT for RL reasoning.}
\citet{liu2025understanding} analyze cold-start RLVR failure modes and motivate
the Dr.~GRPO loss correction we adopt; \citet{deepseek2025r1} demonstrate
strong reasoning via RLVR but rely on FFT throughout.
To our knowledge, no prior work has conducted a systematic comparison of
structured high-rank adapters against LoRA and FFT in the same RLVR
pipeline, nor provided the spectral mechanistic analysis we present here.


\section{Methodology}
\label{sec:methodology}

\subsection{Base Models and Pre-training Scale}
\label{sec:methodology:models}

We study three dense transformer models spanning a wide range of pre-training
scale (Table~\ref{tab:models}), using the \emph{base} checkpoint throughout.
We deliberately use Base-to-RL training to isolate the interaction between
pre-trained model structure, the RL signal, and adaptation parameterization.
An intermediate SFT stage would alter the policy initialization and the weight
state from which RL begins, introducing an additional intervention into the
optimization dynamics we aim to characterize
\citep{liu2025understanding,schulman2025lora}.

\begin{table}[h]
  \centering
  \caption{
    \textbf{Base models evaluated.}
    All models are dense transformers at the 8B parameter scale; pre-training
    token counts are taken from the respective technical reports.
  }
  \label{tab:models}
  \smallskip
  \begin{tabular}{llcc}
    \toprule
    \textbf{Model} & \textbf{Family} & \textbf{Pre-training Scale} & \textbf{Vocab Size} \\
    \midrule
    Apertus 8B     & Independent & ${\sim}15\text{T}$ tokens   & 131K \\
    Llama~3.1~8B   & Meta        & ${\sim}15.6\text{T}$ tokens & 128K \\
    Qwen~3~8B      & Alibaba     & ${\sim}36\text{T}$ tokens   & 150K \\
    \bottomrule
  \end{tabular}
\end{table}

\subsection{Adapter Configurations}
\label{sec:methodology:adapters}

Table~\ref{tab:adapters} summarizes all configurations evaluated, applied to the
same set of linear projection modules in \texttt{bfloat16} throughout.
The LoRA and DoRA rank sweeps are conducted on Qwen~3~8B only; FFT is similarly
restricted to Qwen~3~8B due to compute constraints.
For all LoRA ranks, we fix $\alpha=32$ and use the standard $\alpha/r$
scaling. This follows common post-training practice
\citep{schulman2025lora}; the $1/r$ factor makes the expected initial adapter
update approximately rank-invariant, providing a natural shared scaling
convention for our controlled rank comparison. We additionally evaluate QuanTA with target-module ablations (excluding
key/value projections and excluding only the value projection); full ablation
results are in Appendix~\ref{app:module_ablation}.

\begin{table}[h]
  \centering
  \caption{
    \textbf{Adapter configurations.}
    Trainable parameter counts reported for Qwen~3~8B.
    ``All linear''$^\dagger$ denotes all attention projection and feed-forward linear
    modules with KV excluded.
  }
  \label{tab:adapters}
  \smallskip
  \setlength{\tabcolsep}{5pt}
  \begin{tabular}{llcc}
    \toprule
    \textbf{Method} & \textbf{Configuration} & \textbf{Target Modules}
      & \textbf{Trainable Params} \\
    \midrule
    LoRA  & $r \in \{1,4,8,16,32,64,128,256\}$
      & All linear & ${\sim}$2.7M--696M  \\
    DoRA  & $r \in \{8,16,32,64\}$
          & All linear & 23M--180M          \\
    \midrule
    QuanTA $d=4$ & $[16,8,8,4]$
          & All linear$^\dagger$ & 7.7M              \\
    QuanTA $d=3$ & $[16,16,16]$
          & All linear$^\dagger$ & 35.3M             \\
    \midrule
    FFT   & Full model & All params & ${\sim}$8B \\
    \bottomrule
  \end{tabular}
\end{table}

\subsection{Benchmark Selection and Data Curation}
\label{sec:methodology:data}

We train and evaluate on five benchmarks; full curation details are in
Appendix~\ref{app:data}.
\textbf{DeepMath-Hard} applies a difficulty filter ($\geq 8.5/10$) to
DeepMath-103K \citep{he2025deepmath}, yielding 5,399 hardest problems concentrated in
abstract mathematics; this is our primary discriminative benchmark. This hard-example construction is also motivated by recent evidence that, under
fixed annotation budgets, GRPO post-training benefits most from examples that
are difficult for the base model rather than easy, medium, or randomly selected
examples~\citep{pikus2025hard}.
\textbf{Skywork-Hard} retains the 6,702 problems from Skywork-OR1
\citep{Yang2024skywork} on which DeepSeek-R1-Distill-Qwen-32B achieves a pass
rate below 19\% (Score $\ge 13$), providing a frontier stress test.
\textbf{Enigmata} \citep{enigmata} is a logic and puzzle reasoning benchmark. We draw 10,000 problems while preserving the dataset’s original difficulty distribution, and evaluate the model on the official published evaluation set.
\textbf{MATH} \citep{hendrycks2021math} serves as a standard
reference benchmark.
We additionally probe out-of-distribution generalization by evaluating
DeepMath-Hard-trained checkpoints on the merged \textbf{AIME 2025--2026} \citep{aops2025aimei,aops2025aimeii,aops2026aimei,aops2026aimeii} competition set (60 problems, Avg@32).

\subsection{Reinforcement Learning Protocol}
\label{sec:methodology:rl}

Unless otherwise stated, we train models with GRPO \citep{Shao2024DeepSeekMath}
using group size $G=8$, KL coefficient $\beta=0$, the Dr.~GRPO token-level loss
\citep{liu2025understanding}, and global batch size 64; vLLM
\citep{kwon2023vllm} accelerates generation via PagedAttention.
The batch-size ablation in Section~\ref{sec:experiments:batch_size} varies the
global batch size over $\{64,128,256\}$ while holding the remaining protocol
fixed, with the key $r\in\{8,64,256\}$ boundary configurations additionally
replicated at seed 43. A separate group-size ablation varies
$G\in\{4,8,16\}$ while holding the rest of the protocol fixed.
The reward is a composite $R_{\text{total}} = R_{\text{acc}} + R_{\text{fmt}}$:
$R_{\text{acc}}$ uses depth-aware \texttt{\textbackslash boxed\{\}} extraction
with \texttt{math\_verify} semantic equivalence (1.0 correct, 0.0 otherwise),
and $R_{\text{fmt}}$ applies shaped penalties for structural degeneration
(empty reasoning chains, repetition, hallucinated turns), bounded below
$R_{\text{acc}}$ to prevent reward hacking.
Full reward weights and hyperparameters are in Appendix~\ref{app:implementation}.

\paragraph{Scheduling and cold-start stability.}
\citet{schulman2025lora} report that constant learning-rate schedules are stable
in large-batch Base-to-RL training.
In our batch-size-64 main protocol, constant schedules cause immediate policy
collapse across all methods (Appendix~\ref{app:lr_collapse}).
A cosine decay schedule with warmup ratio 0.1 and minimum learning-rate ratio
0.15 eliminates this failure mode in the batch-size-64 main protocol, so we
use it as a fixed component of all experiments in this paper. We do not claim that this schedule is universally optimal across batch sizes or
update budgets.


\section{Experiments and Results}
\label{sec:experiments}


\subsection{Rank Collapse: The LoRA Rank Sweep in RLVR}
\label{sec:experiments:collapse}

Table~\ref{tab:rank_sweep} presents a systematic LoRA rank sweep on
Qwen~3~8B trained on DeepMath-Hard at global batch size 64.
Performance remains broadly stable across $r=1$--$32$, declines to 73.1\% at
$r=64$, and then collapses abruptly to 4.7\% at $r=128$ and 2.3\% at
$r=256$.
This is qualitatively distinct from the SFT literature's prediction of
diminishing returns: under this RLVR protocol, performance does not merely
plateau at high rank but falls to near-floor accuracy. Section \ref{sec:experiments:spectra} provides a
spectral analysis showing that these collapsed runs concentrate update energy
into a small number of singular directions.

\paragraph{From rank threshold to stability boundary.}
The batch-size ablation in Section~\ref{sec:experiments:batch_size} refines this
finding.
At batch size 64, the observed collapse boundary lies between $r=64$ and
$r=128$.
However, this threshold is not a property of nominal rank alone: under the same
one-epoch GRPO protocol, increasing global batch size to 128 or 256 does not
rescue high-rank LoRA under our fixed protocol and instead moves the observed safe-rank boundary
downward.
Thus, rank collapse should be understood as a rank--batch--optimization
stability boundary for unconstrained LoRA, rather than as a fixed universal
rank cutoff.

\begin{table}[h]
  \centering
  \caption{
  \textbf{LoRA rank sweep on Qwen~3~8B, DeepMath-Hard
  (seed 42, batch size 64).}
  Performance is broadly stable at low rank before an abrupt collapse between
  $r=64$ and $r=128$ under the common training protocol.
}
  \label{tab:rank_sweep}
  \smallskip
  \begin{tabular}{lccc}
    \toprule
    \textbf{Method} & \textbf{Nominal Rank} & \textbf{Trainable Params}
      & \textbf{DeepMath-Hard} \\
    \midrule
    LoRA & $r=1$   & 2.7M  & 79.29\% \\
    LoRA & $r=4$   & 10.9M & 79.29\% \\
    LoRA & $r=8$   & 22M    & 78.1\% \\
    LoRA & $r=16$  & 43.6M  & 76.2\% \\
    LoRA & $r=32$  & 87.2M  & 77.3\% \\
    LoRA & $r=64$  & 174M   & 73.1\% \\
    LoRA & $r=128$ & 348M   & \phantom{0}4.7\% \\
    LoRA & $r=256$ & 696M   & \phantom{0}2.3\% \\
    \bottomrule
  \end{tabular}
\end{table}

\subsection{Batch and Group Scaling Do Not Rescue High-Rank LoRA Under Our Fixed Protocol}
\label{sec:experiments:batch_size}

Two sampling-based explanations for rank collapse are natural.
First, a small global batch may produce noisy optimization updates, suggesting
that larger batches should stabilize high-rank LoRA.
Second, GRPO estimates relative advantages from multiple completions of the
same prompt. Intuitively, increasing the group size $G$ gives a capable base
model more opportunities to produce both successful and unsuccessful
trajectories, potentially yielding a more informative relative-reward signal.
We test both possibilities while keeping the remaining training protocol fixed.

\paragraph{Batch-size scaling.}
We increase global batch size from 64 to 128 and 256 while preserving the same
one-epoch cold-start GRPO protocol, learning-rate schedule, reward function,
group size, and optimizer.
Because DeepMath-Hard contains 5,399 training examples, one epoch corresponds
to 674, 337, and 168 optimizer steps at batch sizes 64, 128, and 256,
respectively.

Panel~(a) of Table~\ref{tab:sampling_scale_ablation} shows that increasing
batch size does not rescue high-rank LoRA.
LoRA $r=128$ and $r=256$ remain near floor across all tested batch sizes.
More strikingly, $r=64$, which reaches 73.1\% at batch size 64, falls to
8.7\% and 6.0\% at batch sizes 128 and 256.
This behavior is not explained by reduced update count alone: under the same
larger-batch regimes, low-rank LoRA and the structured adapters continue to
learn, with QuanTA $d=3$ retaining 65.8\% even at batch size 256.
Thus, larger batches do not eliminate rank collapse under the fixed-pass
protocol; instead, the observed LoRA stability boundary moves downward.

\begin{table}[t]
  \centering
  \caption{
    \textbf{Sampling-scale ablations on Qwen~3~8B, DeepMath-Hard (seed 42).}
    \textbf{(a)} Increasing global batch size does not rescue high-rank LoRA;
    under the fixed one-epoch protocol, larger batches also reduce the number
    of optimizer steps.
    \textbf{(b)} Increasing GRPO group size can improve some configurations,
    but LoRA $r=256$ remains in the collapsed regime throughout.
  }
  \label{tab:sampling_scale_ablation}

  \small

  \begin{minipage}[t]{0.49\linewidth}
    \centering
    \textbf{(a) Global batch-size scaling}
    \vspace{3pt}

    \resizebox{\linewidth}{!}{%
      \begin{tabular}{llccc}
        \toprule
        \textbf{Method} & \textbf{Config}
          & \textbf{BS=64}
          & \textbf{BS=128}
          & \textbf{BS=256} \\
        & & \textbf{674 steps}
          & \textbf{337 steps}
          & \textbf{168 steps} \\
        \midrule
        LoRA   & $r=8$   & 78.1\% & 64.0\% & 20.3\% \\
        LoRA   & $r=64$  & 73.1\% &  8.7\% &  6.0\% \\
        LoRA   & $r=128$ &  4.7\% &  4.1\% &  3.3\% \\
        LoRA   & $r=256$ &  2.3\% &  3.2\% &  3.5\% \\
        \midrule
        DoRA   & $r=16$  & 84.0\% & 67.6\% & 22.5\% \\
        QuanTA & $d=3$   & 84.0\% & 72.2\% & 65.8\% \\
        \bottomrule
      \end{tabular}%
    }
  \end{minipage}
  \hfill
  \begin{minipage}[t]{0.49\linewidth}
    \centering
    \textbf{(b) GRPO group-size scaling}
    \vspace{3pt}

    \resizebox{\linewidth}{!}{%
      \begin{tabular}{llccc}
        \toprule
        \textbf{Method} & \textbf{Config}
          & $\mathbf{G=4}$
          & $\mathbf{G=8}$
          & $\mathbf{G=16}$ \\
        \midrule
        LoRA   & $r=8$   & 75.0\% & 78.1\% & 78.7\% \\
        LoRA   & $r=256$ &  2.3\% &  2.3\% &  6.6\% \\
        FFT    & ---     & 78.1\% & 77.3\% & 79.1\% \\
        QuanTA & $d=3$   & 75.0\% & 84.0\% & 78.5\% \\
        \bottomrule
      \end{tabular}%
    }
  \end{minipage}
\end{table}

\paragraph{Group-size scaling.}
Panel~(b) gives a complementary result.
Increasing $G$ can help some configurations: LoRA $r=8$ improves from 75.0\%
at $G=4$ to 78.7\% at $G=16$, consistent with additional rollouts providing
more opportunities for informative within-group reward variation.
The effect is neither monotonic nor universal, however.
QuanTA peaks at $G=8$, FFT changes little, and LoRA $r=256$ improves only
from 2.3\% to 6.6\% as $G$ increases from 4 to 16, remaining near floor.
Thus, group size affects optimization behavior but does not explain the
qualitative rank-collapse boundary within the tested range.

\paragraph{Replication across seeds.}
We additionally replicate the key LoRA rank--batch configurations with seed
43. The qualitative boundary is nearly unchanged: $r=64$ is functional at
batch size 64 in both seeds (73.1\%/73.8\%) but collapses at batch size 128
(8.7\%/8.6\%) and 256 (6.0\%/4.2\%), while $r=256$ remains near floor
throughout.
The observed rank--batch stability boundary is therefore not specific to
seed 42.

Taken together, the two ablations show that sampling scale can affect
optimization quality but does not by itself remove the high-rank LoRA failure
regime.
The observed boundary instead depends on the interaction between rank,
sampling configuration, optimization horizon, and update parameterization.

Train-time diagnostics at batch size 256 further distinguish slow learning
from non-ignition (Appendix~\ref{app:bs256_dynamics}).
LoRA $r=8$ and DoRA $r=16$ improve steadily but remain update-limited after
168 optimizer steps, whereas LoRA $r=64$, $r=128$, and $r=256$ remain near
floor throughout training.
QuanTA $d=3$ enters the learning regime early and reaches 65.8\%, providing
additional evidence that update structure affects which configurations enter a
productive optimization regime.

These results also refine the comparison to large-batch LoRA reports.
\citet{schulman2025lora} report no rank collapse at $r=128$ using the same
base model.
Our ablations show that neither increasing global batch size nor moderately
increasing GRPO group size is, by itself, sufficient to eliminate high-rank
LoRA collapse under our one-pass cold-start protocol.
We therefore interpret the discrepancy as evidence that rank collapse depends
on the joint optimization regime---including rank, sampling configuration,
update horizon, learning-rate schedule, and parameterization---rather than on
a universal rank threshold.

\subsection{Main Comparison: LoRA, DoRA, QuanTA, and Full Fine-Tuning}
\label{sec:experiments:main}

Table~\ref{tab:main_results} presents results for all primary methods on
Qwen~3~8B (seed 42); multi-seed DeepMath-Hard and Skywork-Hard results are in
Appendix~\ref{app:multiseed}.

\begin{table}[h]
  \centering
  \caption{
    \textbf{Main results on Qwen~3~8B (seed 42).}
    $^\dagger$QuanTA $d=3$ uses the \texttt{no\_kv} target-module configuration;
    see Appendix~\ref{app:module_ablation} for full ablation.
    Best result per benchmark is \textbf{bolded}.
  }
  \label{tab:main_results}
  \smallskip
  \setlength{\tabcolsep}{5pt}
  \begin{tabular}{llcccc}
    \toprule
    \textbf{Method} & \textbf{Config}
      & \textbf{Params}
      & \textbf{MATH}
      & \textbf{DeepMath-Hard}
      & \textbf{Skywork-Hard} \\
    \midrule
    LoRA & $r=8$  & 22M       & \textbf{78.9\%} & 78.1\%          & 19.0\%  \\
    LoRA & $r=16$ & 43.6M     & 78.1\%          & 76.2\%          & 15.2\%  \\
    \midrule
    DoRA & $r=8$  & 23.2M     & 78.1\%          & 81.9\%          & 16.9\%  \\
    DoRA & $r=16$ & 45.0M     & 75.6\%          & \textbf{84.0\%} & 17.0\%  \\
    \midrule
    QuanTA$^\dagger$ & $d=3$  & 35.3M & \textbf{78.9\%} & \textbf{84.0\%} & \textbf{19.4\%} \\
    \midrule
    FFT  & ---    & ${\sim}$8B & \textbf{78.9\%} & 77.3\%          & 15.9\%  \\
    \bottomrule
  \end{tabular}
\end{table}

Three findings stand out.
\textbf{(i)} In seed 42, both QuanTA $d=3$ and DoRA $r=16$ reach 84.0\% on
DeepMath-Hard, compared with 77.3\% for FFT, while updating less than 0.6\% of
the model parameters. Across seeds 42--44, all three methods exhibit run-to-run variation: FFT reaches $73.16\%\pm7.55\%$, DoRA
$r=16$ reaches $74.4\%\pm10.9\%$, and QuanTA $d=3$ reaches
$75.1\%\pm9.6\%$. We therefore interpret the primary result as structured parameter-efficient competitiveness with FFT.

\textbf{(ii)} Trainable parameter count alone does not predict performance.
In seed 42, FFT updates the full ${\sim}8$B parameters yet scores 77.3\% on
DeepMath-Hard, compared with 78.1\% for LoRA $r=8$ and 84.0\% for both
structured adapters. Section~\ref{sec:experiments:spectra} examines the
corresponding differences in learned update geometry.
On MATH, all methods saturate near 78.9\% with the exception of DoRA $r=16$
(75.6\%); given that multi-seed results are unavailable for this configuration
on MATH, we do not interpret this single-seed difference further.

\textbf{(iii)} The sampling-scale ablations show that the structured-adapter
advantage in optimization stability is not limited to the original training
configuration. At batch sizes 128 and 256, QuanTA $d=3$ remains trainable while
higher-rank vanilla LoRA enters the near-floor regime; at batch size 256,
QuanTA reaches 65.8\% after 168 optimizer steps while LoRA $r=64$, $r=128$,
and $r=256$ remain near floor. Varying GRPO group size likewise changes
performance without rescuing LoRA $r=256$. Together, these results support the
view that structured parameterizations occupy a wider stable optimization
region under the tested cold-start GRPO protocol.

Full multi-seed results are in
Appendix~\ref{app:multiseed}. The AIME evaluation provides a separate,
suggestive out-of-distribution probe and is reported in
Appendix~\ref{app:aime}.

\subsection{The Model Maturity Hypothesis}
\label{sec:experiments:maturity}

Table~\ref{tab:maturity_teaser} presents results across all three base models.
Both ${\sim}15$T models score near floor on DeepMath-Hard under any adapter
configuration; Qwen~3~8B reaches 78--84\%.
The behavioral contrast extends beyond absolute accuracy: on MATH,
Llama~3.1~8B exhibits seed-level instability under QuanTA $d=4$ absent from
LoRA---seed 43 collapses entirely (0.8\%), while LoRA $r=16$ is stable across
all three seeds (30.2\%~$\pm$~3.7\%); full multi-seed results are in
Appendix~\ref{app:multiseed}.
Qwen~3~8B shows no such instability under any structured adapter on any benchmark.

\begin{wrapfigure}{l}{0.54\textwidth}
  \vspace{-0.7\baselineskip}
  \centering
  \includegraphics[width=\linewidth]{
    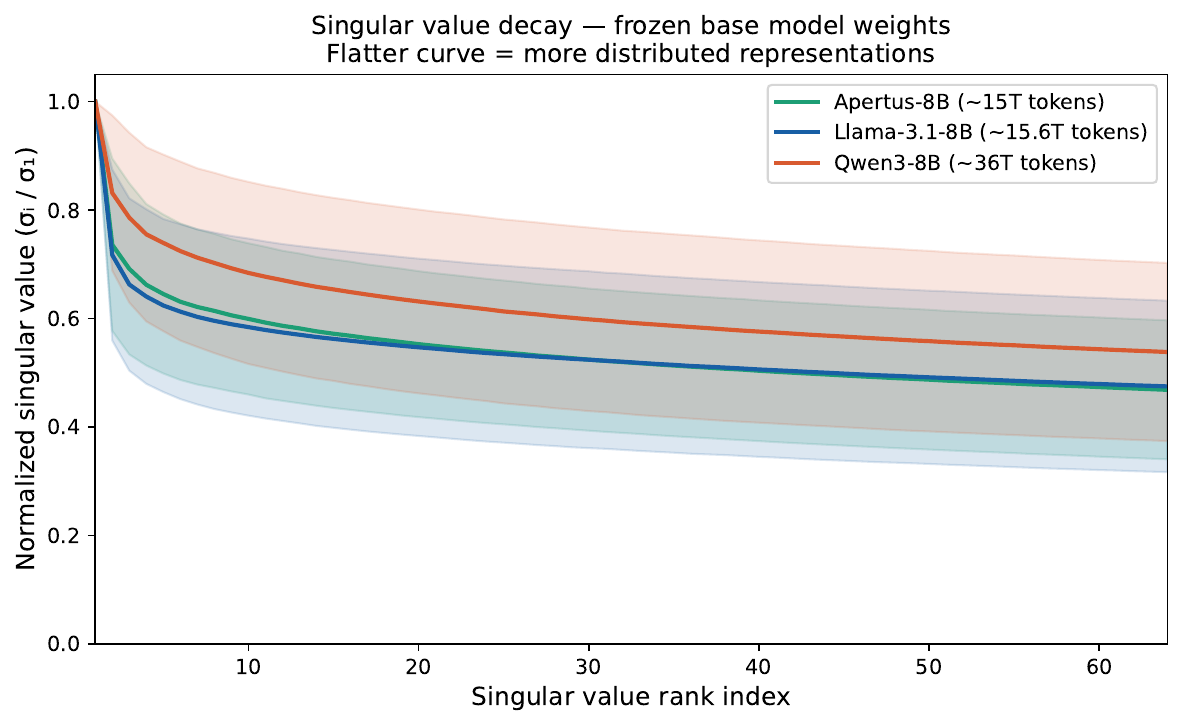
  }
  \caption{
    \textbf{Frozen-weight singular-value decay.}
    Top 64 singular values averaged across sampled linear layers;
    shading denotes $\pm 1$ standard deviation.
    Qwen~3~8B exhibits markedly flatter spectra than the two
    ${\sim}15$T models.
  }
  \label{fig:frozen_decay_main}
  \vspace{-0.6\baselineskip}
\end{wrapfigure}

Figure~\ref{fig:frozen_decay_main} offers a structural correlate that precedes
any fine-tuning. The singular value decay of frozen linear layers reveals a
monotonic ordering: both ${\sim}15$T models decay steeply and nearly identically,
while Qwen~3~8B's curve sits markedly flatter throughout. This pattern is
consistent with pre-training data volume being an important driver, although
architecture, tokenizer, and data mixture remain confounded. We hypothesize
that flatter singular value spectra indicate more distributed pre-trained
representations, making adaptation benefit from updates spread across many
singular directions. This interpretation is consistent with QuanTA's advantage
on Qwen~3~8B, but controlled validation with matched architectures and
intermediate pre-training checkpoints is needed to establish causality.
More details can be found in Appendix~\ref{app:frozen_spectra}.

\subsection{Spectral Analysis: Update Geometry Beyond Nominal Rank}
\label{sec:experiments:spectra}

\begin{figure}[h]
  \centering
  \includegraphics[width=\linewidth]{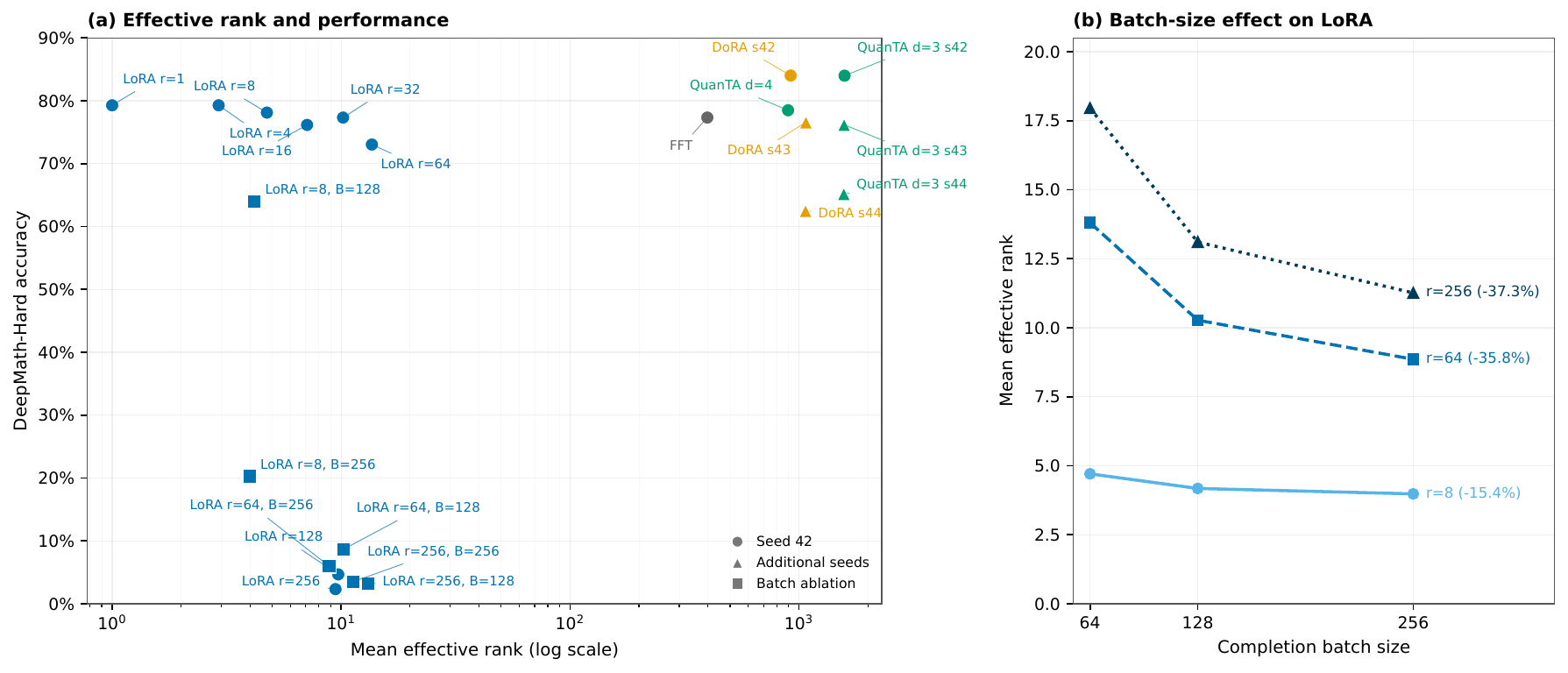}
  \caption{
    \textbf{Effective rank, performance, and batch-size dependence on
    Qwen~3~8B, DeepMath-Hard.}
    \textbf{(a)} Mean effective rank versus accuracy across LoRA ranks,
    structured adapters, FFT, additional seeds, and LoRA batch-size ablations.
    Colors distinguish methods; circles denote the original seed-42 runs,
    triangles additional seeds, and squares batch-size ablations.
    The low-rank LoRA regime occupies a compact low-effective-rank region,
    while structured adapters occupy a substantially higher-effective-rank
    regime. At the same time, variation within and across methods shows that
    effective rank alone does not determine accuracy.
    \textbf{(b)} Mean effective rank of LoRA $r\in\{8,64,256\}$ as global
    batch size increases from 64 to 256. Effective rank decreases for all
    three ranks, with the largest relative reductions at $r=64$ and $r=256$.
  }
  \label{fig:effective_rank}
\end{figure}

Figure~\ref{fig:effective_rank} extends the spectral analysis across the full
LoRA rank sweep, the structured adapters, additional seeds, and the LoRA
batch-size ablation.
For each learned update $\Delta W$, we compute the effective rank as the
participation ratio
$\rho = \exp(H(p))$,
where $p$ is the normalized distribution of squared singular values and
$H$ is Shannon entropy, and then average $\rho$ across target layers.

Three observations emerge.

\textbf{(i) The low-rank LoRA regime is spectrally coherent, but the
high-rank failure cannot be reduced to a single effective-rank threshold.}
The newly added LoRA $r=1$ and $r=4$ checkpoints extend the same qualitative
pattern seen in Section~\ref{sec:experiments:collapse}: the low-rank
regime combines strong performance with compact, low-dimensional updates.
As nominal rank increases from $r=1$ through $r=64$, mean effective rank
increases gradually while performance remains broadly stable before declining.
However, the collapsed $r=128$ and $r=256$ checkpoints do not simply have the
lowest effective rank.
In particular, at batch size 64, LoRA $r=8$, $r=64$, and $r=256$ achieve mean
effective ranks of 4.709, 13.807, and 17.956, respectively.
Thus, increasing nominal rank by $32\times$ from $r=8$ to $r=256$ produces
only a modest increase in effective rank, and the collapsed $r=256$ checkpoint
still has higher effective rank than the functional $r=64$ checkpoint.
Rank collapse therefore cannot be explained by effective rank alone.
Instead, the spectral evidence shows that increasing nominal rank does not
translate into proportionally broader usable updates.

\textbf{(ii) Structured adapters occupy a qualitatively different spectral
regime from both LoRA and FFT.}
Across seeds, DoRA and QuanTA consistently produce substantially
higher-effective-rank updates than vanilla LoRA, despite updating less than
0.6\% of the model parameters.
FFT occupies an intermediate spectral regime: although it updates the full
${\sim}8$B parameters, its learned update remains markedly lower in effective
rank than either structured adapter.
The additional-seed points preserve this clustering while also making visible
the corresponding run-to-run performance variation.
Taken together, these results show that trainable parameter count alone does
not determine how broadly an optimization method distributes its update.

\textbf{(iii) Batch scaling systematically contracts LoRA's learned
spectrum.}
Panel~(b) isolates the effect of global batch size on LoRA.
As batch size increases from 64 to 256, mean effective rank decreases from
$4.709 \rightarrow 4.180 \rightarrow 3.983$ for $r=8$,
$13.807 \rightarrow 10.275 \rightarrow 8.861$ for $r=64$, and
$17.956 \rightarrow 13.106 \rightarrow 11.264$ for $r=256$,
corresponding to relative reductions of 15.4\%, 35.8\%, and 37.3\%,
respectively.
This spectral contraction accompanies very different performance responses:
$r=8$ remains trainable but degrades, $r=64$ moves from the functional regime
to the collapsed regime, and $r=256$ is already collapsed at batch size 64.
The relationship between effective rank and policy quality is therefore
parameterization- and optimization-dependent rather than monotonic.

Overall, Figure~\ref{fig:effective_rank} refines the central claim of the
paper.
Effective rank is a useful diagnostic of how optimization distributes an
update, but it is not itself the operative scalar variable.
The more informative organizing principle is update geometry:
structured parameterizations produce substantially more distributed learned
updates while constraining the set of solutions reachable by gradient descent,
whereas nominal rank and raw trainable parameter count alone provide limited
information about the geometry ultimately realized by RL optimization.

\subsection{Exploratory Evaluation Beyond Mathematics: Enigmata}
\label{sec:experiments:enigmata}

To test whether the qualitative adapter behavior observed on mathematical
reasoning can also appear in a different reasoning domain, we train the primary
methods on Enigmata 10K under the same cold-start GRPO protocol and evaluate on
the disjoint official Enigmata evaluation set (Table~\ref{tab:enigmata}).

\begin{table}[h]
  \centering
  \caption{
    \textbf{Enigmata results (Qwen~3~8B).}
    Models are trained on Enigmata 10K and evaluated on the disjoint official
    Enigmata evaluation set (pass@1).
  }
  \label{tab:enigmata}
  \smallskip
  \setlength{\tabcolsep}{6pt}
  \begin{tabular}{llcc}
    \toprule
    \textbf{Method} & \textbf{Config}
      & \textbf{Enigmata 10K (train-time)}
      & \textbf{Enigmata Eval (pass@1)} \\
    \midrule
    LoRA   & $r=8$              & 3.3\%  & 11.8\% \\
    DoRA   & $r=16$             & 7.7\%  & 16.6\% \\
    FFT    & ---                & 9.8\%  & 19.1\% \\
    QuanTA & $d=3$ (\texttt{no\_kv}) & 12.7\% & \textbf{20.6\%} \\
    \bottomrule
  \end{tabular}
\end{table}

The same qualitative ordering appears in both training-time and held-out
evaluation signals:
QuanTA (20.6\%) $>$ FFT (19.1\%) $>$ DoRA (16.6\%) $>$ LoRA (11.8\%).
This provides an exploratory indication that the relative behavior of the
tested parameterizations is not specific to the mathematical datasets used in
our main experiments.
In particular, the structured adapters again improve over vanilla LoRA, while
QuanTA remains competitive with FFT using a small fraction of the trainable
parameters.

Because this experiment uses a single seed on one non-mathematical reasoning
domain, we do not interpret the precise ordering as a general ranking of the
methods.
Rather, Enigmata serves as a complementary probe showing that the qualitative
structure-sensitive behavior identified on mathematical RLVR can also arise on
logic and puzzle reasoning.



\section{Discussion and Limitations}
\label{sec:discussion}

\paragraph{Interpreting rank collapse.}
The ablations narrow several simple explanations for rank collapse.
Increasing global batch size reduces per-update sampling noise but does not
rescue high-rank LoRA under our fixed-pass protocol; instead, the observed
stability boundary moves downward, with $r=64$ failing at batch sizes 128 and
256 in both tested seeds.
Increasing GRPO group size provides a complementary test.
More completions per prompt can improve some configurations---LoRA $r=8$
benefits modestly as $G$ increases---but $r=256$ remains near floor throughout
$G\in\{4,8,16\}$.
Thus, neither a small global batch nor the original choice of $G=8$ alone
explains the qualitative failure regime.

The spectral results further refine the mechanism.
High-rank LoRA uses only a small fraction of its nominal update dimensionality,
but collapse is not determined by a scalar effective-rank threshold: the
collapsed $r=256$ checkpoint can have higher effective rank than a functional
$r=64$ checkpoint.
Likewise, increasing batch size systematically contracts LoRA's effective rank
while producing rank-dependent performance responses.
By contrast, DoRA and QuanTA consistently occupy a substantially more
distributed spectral regime and remain stable over a wider tested optimization
region, while FFT achieves lower effective rank despite updating the full
model.
We therefore view effective rank as a useful diagnostic of learned update structure,
not a predictor of policy quality.
The broader organizing principle is the geometry induced by the
parameterization: different parameterizations constrain which updates are
reachable by gradient descent even when their nominal capacity is large.

\paragraph{Model Maturity Hypothesis: evidence and limits.}
Two independent observations motivate the hypothesis.
Behaviorally, Apertus and Llama~3.1---different model families with similar
reported pre-training scale---exhibit closely related adapter behavior, while
Qwen~3~8B occupies a qualitatively different regime.
Spectrally, the same partition is visible before fine-tuning: the two
${\sim}15$T-token models have closely overlapping frozen-weight singular-value
decay, whereas the ${\sim}36$T-token Qwen~3 model is markedly flatter
(Appendix~\ref{app:frozen_spectra}).
Among the model-level quantities measured in our comparison, reported
pre-training scale aligns most directly with both partitions.
Architecture, tokenizer, and data mixture nevertheless co-vary across models,
so the present evidence motivates a falsifiable relationship rather than
causally isolating pre-training scale.

\paragraph{Practical takeaways.}
(1)~\textbf{Treat rank as an optimization variable, not simply a capacity
knob.} Under a shared cold-start GRPO recipe, increasing LoRA rank can move the
policy abruptly from a functional to a collapsed regime rather than merely
yielding diminishing returns.
(2)~\textbf{More sampling is not a universal stabilizer.} Larger global
batches and larger GRPO groups can change performance, but neither reliably
rescues collapsed high-rank LoRA in the regimes tested here.
(3)~\textbf{Structure offers a third option beyond the FFT--LoRA binary.}
DoRA and QuanTA remain competitive with FFT on the primary multi-seed
comparison while updating less than 0.6\% of the model parameters and occupying
a wider tested stability region than high-rank vanilla LoRA.
(4)~\textbf{Optimization details remain consequential.} In our batch-size-64
cold-start setting, cosine decay with warmup avoids the schedule-level collapse
observed with a constant learning rate (Appendix~\ref{app:lr_collapse}).

\paragraph{Limitations.}
\begin{itemize}

  \item \textbf{Model scale and architecture.}
  Our experiments use 8B dense transformers; whether the same stability
  boundaries arise in substantially larger models or MoE architectures remains
  open.

  \item \textbf{Cold-start scope.}
  We deliberately use Base-to-RL training to isolate the interaction between
  pre-trained model structure, RL feedback, and adaptation parameterization.
  An intermediate SFT stage changes the policy and weight initialization from
  which RL begins, so the boundaries measured here need not transfer unchanged
  to SFT-to-RL pipelines.

  \item \textbf{Controlled protocol rather than per-configuration optimum.}
  The rank comparisons hold the LoRA scaling convention, learning rate,
  scheduler, optimizer, and one-epoch training budget fixed in order to study
  relative optimization behavior under a shared recipe.
  Rank-specific tuning could shift the observed boundaries. We do not test
batch sizes below 64 or claim that 64 is optimal.

  \item \textbf{Maturity and domain evidence.}
  The three-model comparison cannot disentangle reported pre-training scale
  from architecture, tokenizer, and data mixture.
  The Enigmata experiment additionally provides one exploratory
  non-mathematical evaluation rather than evidence of broad domain
  generalization.

\end{itemize}


\section{Conclusion}
\label{sec:conclusion}

RLVR adaptation is commonly framed as a trade-off between the efficiency of
LoRA and the expressivity of Full Fine-Tuning.
Our results suggest that this framing omits an important variable: the geometry
of the update induced by the parameterization.
Under a common cold-start GRPO protocol, LoRA performance remains broadly
stable across low ranks before undergoing an abrupt high-rank collapse.
The qualitative boundary reproduces across seeds, and neither increasing global
batch size nor increasing GRPO group size within the tested ranges restores the
collapsed high-rank regime.

More broadly, our results suggest that nominal rank and trainable parameter
count are incomplete descriptions of adaptation capacity in RLVR.
Structured parameterizations can reach qualitatively different update
geometries while remaining highly parameter-efficient, making the relevant
question not simply how many parameters to update, but what structure to impose
on the updates that optimization can reach.

The Model Maturity Hypothesis extends this view to the base model itself:
the most suitable update geometry may depend on structure already acquired
during pre-training. This motivates a broader direction for efficient RLVR in
which adaptation is chosen jointly with the model and optimization regime,
rather than through a fixed FFT-versus-LoRA rule.

\section*{Acknowledgments}

This work was supported in part by development access to the MareNostrum 5 supercomputer through the EuroHPC Joint Undertaking. We also acknowledge support from NVIDIA through an Academic Grant that provided additional computational resources for the experiments in this study.

\bibliographystyle{plainnat}
\bibliography{main}

@misc{pikus2025hard,
      title={Hard Examples Are All You Need: Maximizing GRPO Post-Training Under Annotation Budgets}, 
      author={Benjamin Pikus and Pratyush Ranjan Tiwari and Burton Ye},
      year={2025},
      eprint={2508.14094},
      archivePrefix={arXiv},
      primaryClass={cs.LG},
      url={https://arxiv.org/abs/2508.14094}, 
}

@article{schulman2025lora,
  author = {John Schulman and Thinking Machines Lab},
  title = {LoRA Without Regret},
  journal = {Thinking Machines Lab: Connectionism},
  year = {2025},
  url = {https://thinkingmachines.ai/blog/lora/},
  doi = {10.64434/tml.20250929},
}

@inproceedings{
    chen2024quanta,
    author = {Chen, Zhuo and Dangovski, Rumen and Loh, Charlotte and Dugan, Owen and Luo, Di and Solja\v{c}i\'{c}, Marin},
    booktitle = {Advances in Neural Information Processing Systems},
    doi = {10.52202/079017-2928},
    editor = {A. Globerson and L. Mackey and D. Belgrave and A. Fan and U. Paquet and J. Tomczak and C. Zhang},
    pages = {92210--92245},
    publisher = {Curran Associates, Inc.},
    title = {{QuanTA}: Efficient High-Rank Fine-Tuning of LLMs with Quantum-Informed Tensor Adaptation},
    url = {https://proceedings.neurips.cc/paper_files/paper/2024/file/a7c17115db36193f6b83b71b0fe1d416-Paper-Conference.pdf},
    volume = {37},
    year = {2024}
}

@misc{Shao2024DeepSeekMath,
      title={DeepSeekMath: Pushing the Limits of Mathematical Reasoning in Open Language Models}, 
      author={Zhihong Shao and Peiyi Wang and Qihao Zhu and Runxin Xu and Junxiao Song and Xiao Bi and Haowei Zhang and Mingchuan Zhang and Y. K. Li and Y. Wu and Daya Guo},
      year={2024},
      eprint={2402.03300},
      archivePrefix={arXiv},
      primaryClass={cs.CL},
      url={https://arxiv.org/abs/2402.03300}, 
}

@misc{Liu2025Understanding,
      title={Understanding R1-Zero-Like Training: A Critical Perspective}, 
      author={Zichen Liu and Changyu Chen and Wenjun Li and Penghui Qi and Tianyu Pang and Chao Du and Wee Sun Lee and Min Lin},
      year={2025},
      eprint={2503.20783},
      archivePrefix={arXiv},
      primaryClass={cs.LG},
      url={https://arxiv.org/abs/2503.20783}, 
}

@misc{Hu2021LoRA,
      title={LoRA: Low-Rank Adaptation of Large Language Models}, 
      author={Edward J. Hu and Yelong Shen and Phillip Wallis and Zeyuan Allen-Zhu and Yuanzhi Li and Shean Wang and Lu Wang and Weizhu Chen},
      year={2021},
      eprint={2106.09685},
      archivePrefix={arXiv},
      primaryClass={cs.CL},
      url={https://arxiv.org/abs/2106.09685}, 
}

@misc{Jiang2024MoRA,
      title={{MoRA}: High-rank updating for parameter-efficient fine-tuning},
      author={Ting Jiang and Shaohan Huang and Shengyue Luo and others},
      year={2024},
      eprint={2405.12130},
      archivePrefix={arXiv},
      url={https://arxiv.org/abs/2405.12130}
}

@misc{Liu2023BOFT,
      title={Parameter-efficient orthogonal finetuning via butterfly factorization},
      author={Weiyang Liu and Zeju Qiu and Yao Feng and others},
      year={2023},
      eprint={2311.06243},
      archivePrefix={arXiv},
      url={https://arxiv.org/abs/2311.06243}
}

@misc{he2025deepmath,
      title={DeepMath-103K: A large-scale, challenging, decontaminated, and verifiable mathematical dataset for advancing reasoning},
      author={Zhiwei He and Yuehua Chen and Ting Liang and others},
      year={2025},
      eprint={2504.11456},
      archivePrefix={arXiv},
      url={https://arxiv.org/abs/2504.11456}
}

@misc{Yang2024skywork,
      title={{Skywork-Math:} Data scaling laws for mathematical reasoning in large language models — the story goes on},
      author={Liang Zeng and Liangjun Zhong and Liang Zhao and others},
      year={2024},
      eprint={2407.08348},
      archivePrefix={arXiv},
      url={https://arxiv.org/abs/2407.08348}
}

@article{Hendrycks2021Math,
  author = {Dan Hendrycks and Collin Burns and Saurav Kadavath and Akul Arora and Steven Basart and Eric Tang and Dawn Song and Jacob Steinhardt},
  title = {Measuring Mathematical Problem Solving With the MATH Dataset},
  journal = {CoRR},
  volume = {abs/2103.03874},
  year = {2021},
  url = {https://arxiv.org/abs/2103.03874},
  eprinttype = {arXiv},
  eprint = {2103.03874}
}

@inproceedings{liu2024dora,
  title     = {{DoRA}: Weight-Decomposed Low-Rank Adaptation},
  author    = {Liu, Shih-Yang and Wang, Chien-Yi and Yin, Hongxu and Molchanov, Pavlo and Wang, Yu-Chiang Frank and Cheng, Kwang-Ting and Chen, Min-Hung},
  booktitle = {Proceedings of the 41st International Conference on Machine Learning},
  year      = {2024},
  publisher = {PMLR},
  volume    = {235},
  pages     = {32100--32121},
  url       = {https://proceedings.mlr.press/v235/liu24bn.html}
}

@misc{aghajanyan2020intrinsic,
      title={Intrinsic dimensionality explains the effectiveness of language model fine-tuning},
      author={Armen Aghajanyan and Luke Zettlemoyer and Sonal Gupta},
      year={2020},
      eprint={2012.13255},
      archivePrefix={arXiv},
      url={https://arxiv.org/abs/2012.13255}
}

@misc{deepseek2025r1,
      title={DeepSeek-R1: Incentivizing Reasoning Capability in LLMs via Reinforcement Learning}, 
      author={DeepSeek-AI},
      year={2025},
      eprint={2501.12948},
      archivePrefix={arXiv},
      primaryClass={cs.CL},
      url={https://arxiv.org/abs/2501.12948}, 
}

@misc{kwon2023vllm,
      title={Efficient Memory Management for Large Language Model Serving with PagedAttention}, 
      author={Woosuk Kwon and Zhuohan Li and Siyuan Zhuang and Ying Sheng and Lianmin Zheng and Cody Hao Yu and Joseph E. Gonzalez and Hao Zhang and Ion Stoica},
      year={2023},
      eprint={2309.06180},
      archivePrefix={arXiv},
      primaryClass={cs.LG},
      url={https://arxiv.org/abs/2309.06180}, 
}

@misc{enigmata,
      title={Enigmata: Scaling Logical Reasoning in Large Language Models with Synthetic Verifiable Puzzles},
      author={Jiangjie Chen and Qianyu He and Siyu Yuan and Aili Chen and Zhicheng Cai and Weinan Dai and Hongli Yu and Qiying Yu and Xuefeng Li and Jiaze Chen and Hao Zhou and Mingxuan Wang},
      year={2025},
      eprint={2505.19914},
      archivePrefix={arXiv},
      primaryClass={cs.CL},
      url={https://arxiv.org/abs/2505.19914},
}

@article{albert2025randlora,
  title   = {RandLoRA: Full-Rank Parameter-Efficient Fine-Tuning of Large Models},
  author  = {Albert, Paul and Zhang, Frederic Z. and Saratchandran, Hemanth and Rodriguez-Opazo, Cristian and van den Hengel, Anton and Abbasnejad, Ehsan},
  journal = {arXiv preprint arXiv:2502.00987},
  year    = {2025}
}

@misc{aops2025aimei,
  title        = {2025 {AIME} {I} Problems},
  author       = {{Art of Problem Solving}},
  year         = {2025},
  howpublished = {\url{https://artofproblemsolving.com/wiki/index.php/2025_AIME_I_Problems}},
  note         = {AoPS Wiki. American Invitational Mathematics Examination I problems and answer key.}
}

@misc{aops2025aimeii,
  title        = {2025 {AIME} {II} Problems},
  author       = {{Art of Problem Solving}},
  year         = {2025},
  howpublished = {\url{https://artofproblemsolving.com/wiki/index.php/2025_AIME_II_Problems}},
  note         = {AoPS Wiki. American Invitational Mathematics Examination II problems and answer key.}
}

@misc{aops2026aimei,
  title        = {2026 {AIME} {I} Problems},
  author       = {{Art of Problem Solving}},
  year         = {2026},
  howpublished = {\url{https://artofproblemsolving.com/wiki/index.php/2026_AIME_I_Problems}},
  note         = {AoPS Wiki. American Invitational Mathematics Examination I problems and answer key.}
}

@misc{aops2026aimeii,
  title        = {2026 {AIME} {II} Problems},
  author       = {{Art of Problem Solving}},
  year         = {2026},
  howpublished = {\url{https://artofproblemsolving.com/wiki/index.php/2026_AIME_II_Problems}},
  note         = {AoPS Wiki. American Invitational Mathematics Examination II problems and answer key.}
}

\newpage
\appendix

\section{Implementation Details}

\label{app:implementation}

To ensure complete reproducibility of our ``Cold Start'' RLVR experiments, we detail the exact computational environment, hyperparameters, and reward definitions used.

\subsection{Infrastructure and Environment}
All experiments were conducted on 4xH100 node. We used the Axolotl training framework with the GRPO trainer from TRL integrated into Axolotl. To maintain consistency across runs, we utilized a custom Singularity container. The precise hardware and software stack specifications are listed in Table \ref{tab:infrastructure}.

\begin{table}[h]
\centering
\caption{\textbf{Computational Infrastructure.}}
\label{tab:infrastructure}
\begin{sc}
\begin{tabular}{ll}
\toprule
\textbf{Component} & \textbf{Specification} \\
\midrule
\textbf{Compute Hardware} & 4$\times$ NVIDIA H100 (64GB) node \\
\textbf{Operating System} & Ubuntu 24.04 LTS (via Singularity) \\
\textbf{CUDA Driver} & Version 12.8.0 \\
\textbf{Precision} & \texttt{bfloat16} (Training \& Inference) \\
\midrule
\textbf{Deep Learning Framework} & PyTorch 2.9.0 (with Custom \texttt{torchao} 0.13.0) \\
\textbf{Training Orchestration} & Axolotl 0.13.0 \\
\textbf{Inference Engine} & vLLM 0.11.1 (PagedAttention enabled) \\
\textbf{Acceleration Kernels} & FlashAttention 2.8.3, Xformers 0.0.33 \\
\textbf{Distributed Strategy} & PyTorch DDP (DistributedDataParallel) \\
\bottomrule
\end{tabular}
\end{sc}
\end{table}

\subsection{Hyperparameter Configuration}
\label{app:hyperparams}

We utilized the Axolotl training framework with GRPO trainer by TRL integrated into Axolotl. Table \ref{tab:training_params} details the global optimization parameters held constant across runs to ensure fair comparison. Table \ref{tab:peft_params} details the specific configurations for the adaptation methods and task-specific adjustments.

\begin{table}[h]
\centering
\caption{
\textbf{Global RL training hyperparameters for the main protocol.}
Unless otherwise stated, experiments use global batch size 64 and $G=8$;
the batch- and group-size ablations vary the corresponding quantity while
holding the remaining protocol fixed.
}
\label{tab:training_params}
\begin{sc}
\begin{tabular}{ll}
\toprule
\textbf{Parameter} & \textbf{Value} \\
\midrule
\textbf{RL Algorithm} & Group Relative Policy Optimization (GRPO) \\
\textbf{Loss Function} & Dr. GRPO (Token-level unbiased) \\
\textbf{Group Size ($G$)} & 8; $\{4,16\}$ in group-size ablation \\
\textbf{KL Coefficient ($\beta$)} & 0.0 (Pure Outcome-Based RL) \\
\textbf{Reward Scaling} & \texttt{False} (Raw scores used) \\
\midrule
\textbf{Global Batch Size} & 64; $\{128,256\}$ in batch-size ablation \\
\textbf{Optimizer} & AdamW (Fused) \\
\textbf{Weight Decay} & 0.0 \\
\textbf{Scheduler} & Cosine Decay \\
\textbf{Warmup Ratio} & 0.1 (10\% of steps) \\
\textbf{Min LR Ratio} & 0.15 \\
\textbf{Num Epochs} & 1 \\
\bottomrule
\end{tabular}
\end{sc}
\end{table}

\begin{table}[h]
\centering
\caption{\textbf{Method-Specific and Task-Specific Configurations.} We highlight the drastic difference in parameter count between LoRA and QuanTA. LoRA uses $\alpha=32$ across the full rank sweep.}
\label{tab:peft_params}
\begin{sc}
\resizebox{\textwidth}{!}{
\begin{tabular}{l|ccc}
\toprule
\textbf{Configuration} & \textbf{LoRA / DoRA ($r=16$)} & \textbf{QuanTA ($d=4$)} & \textbf{QuanTA ($d=3$)} \\
\midrule
\textbf{Rank / Dims} & $r=16, \alpha=32$ & $d=4, \text{feats}=[16,8,8,4]$ & $d=3, \text{feats}=[16,16,16]$ \\
\textbf{Target Modules} & All Linear & All Linear & All Linear \\
\textbf{Dropout} & 0.0 & 0.0 & 0.0 \\
\textbf{Bias} & None & None & None \\
\textbf{Trainable Params (LLAMA 3.1 8B)} & 41.9M & 6.9M & N/A \\
\textbf{Trainable Params (Qwen 3 8B)} & 43.6M & 7.7M & 35.3M \\
\midrule
\multicolumn{4}{l}{\textit{Task-Specific Learning Rates \& Context}} \\
\midrule
\textbf{MATH / DeepMath / Skywork} & \multicolumn{3}{c}{LR: $1\times 10^{-5}$ \quad | \quad Max Length: 8,192} \\
\bottomrule
\end{tabular}
}
\end{sc}
\end{table}

\subsection{Reward Definitions and Verification}
\label{app:rewards}

In the ``Cold Start'' regime, the model must simultaneously learn formatting (XML tags) and reasoning logic. We utilized a composite reward function $R_{total} = R_{format} + R_{accuracy}$ with equal weighting ($w=1.0$).

\subsubsection{Accuracy Reward ($R_{accuracy}$)}
To evaluate correctness, we strip all formatting tags (e.g., \texttt{<think>}) and extract the final answer using a hierarchical strategy:
\begin{enumerate}
    \item \textbf{Robust Box Extraction:} We scan for the \textbf{last} occurrence of \texttt{\textbackslash boxed\{...\}}, utilizing depth-aware parsing to correctly capture mathematical expressions containing nested braces.
    \item \textbf{Semantic Verification:} We utilize the \texttt{math\_verify} library to compare the extracted answer against the ground truth. This handles symbolic equivalences (e.g., $\frac{1}{2} = 0.5$, $x+y = y+x$) that strict string matching would miss.
    \item \textbf{Score:} Returns \textbf{1.0} if semantically equivalent, \textbf{0.0} otherwise.
\end{enumerate}

\subsubsection{Format Shaping Reward ($R_{\text{fmt}}$)}
To induce Chain-of-Thought reasoning without Supervised Fine-Tuning, we impose structural constraints to penalize degeneration (e.g., empty thoughts, infinite loops). The shaping reward is computed as a weighted sum of satisfied constraints:

\begin{equation}
    R_{\text{fmt}}(y) = \sum_{i} w_i \cdot \mathbb{I}(C_i(y))
\end{equation}

\textbf{Constraints ($C_i$) and Weights ($w_i$):}
\begin{itemize}
    \item \textbf{Structure ($+0.375$):} The output contains a valid closing \texttt{</think>} tag.
    \item \textbf{Answer Existence ($+0.375$):} Non-empty content exists after the reasoning block.
    \item \textbf{Hallucination ($-0.75$):} The model generates simulated user turns (e.g., ``User:'').
    \item \textbf{Lazy Thinking ($-0.6$):} The content inside \texttt{<think>...</think>} is empty or null.
    \item \textbf{Repetition ($-0.5$):} Line-level n-gram repetition exceeds the degeneration threshold.
    \item \textbf{Malformed XML ($-0.3$):} Tags are missing or unclosed.
\end{itemize}

\textit{Note: The maximum cumulative format bonus ($0.75$) is bounded below the accuracy reward ($1.0$) to prevent reward hacking, ensuring the optimizer prioritizes correctness over mere template compliance.}

\section{Data Curation \& ``Hardcore'' Filtering Strategy}
\label{app:data}

\subsection{DeepMath-Hard Construction}
We utilized the \texttt{DeepMath-103K} dataset \citep{he2025deepmath}, which aggregates high-difficulty problems from sources like NuminaMath and MathStackExchange. The original dataset contains a difficulty metadata field ranging from 0.0 to 10.0.

\begin{figure}[h]
    \centering
    \includegraphics[width=0.8\linewidth]{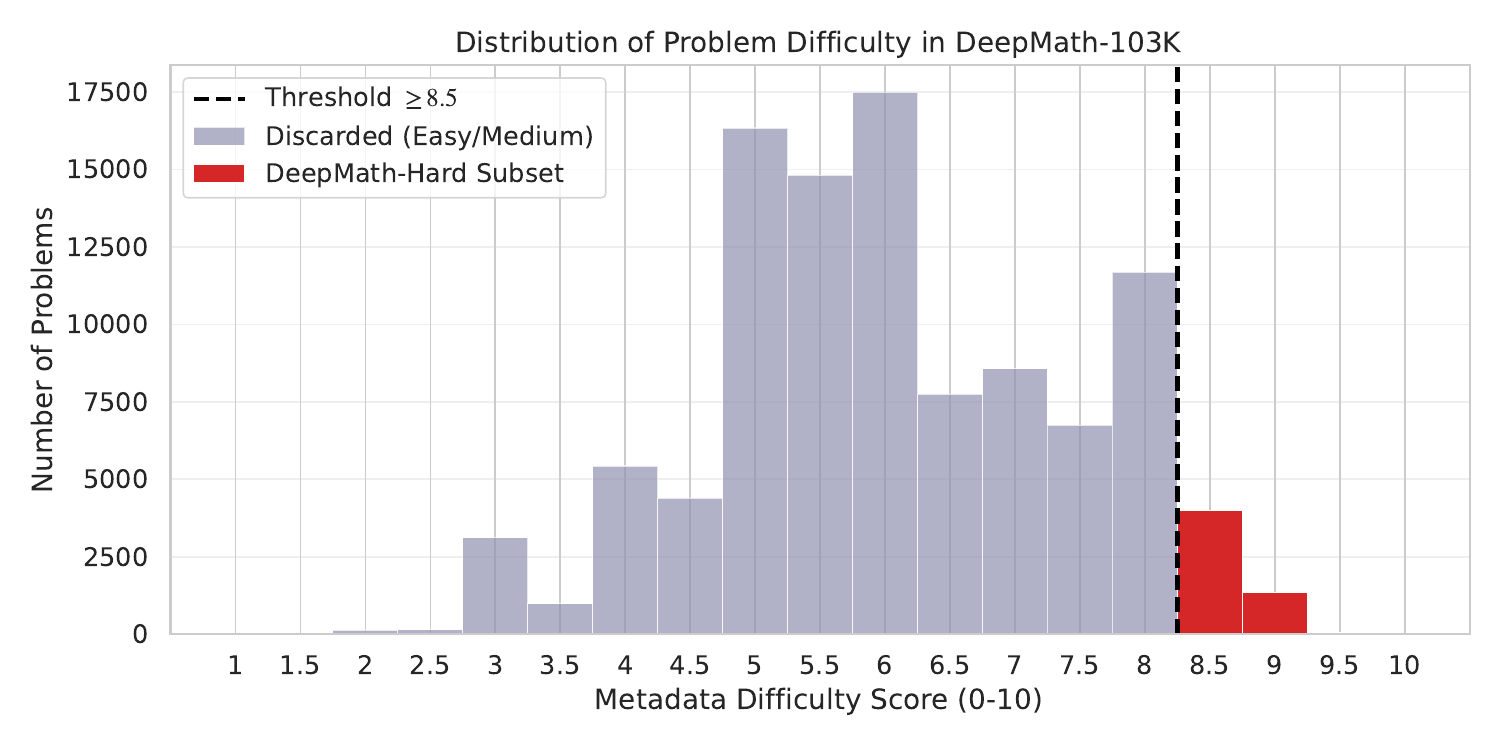}
    \caption{\textbf{Difficulty Distribution of DeepMath-103K.} The dashed line represents our cut-off at difficulty score 8.5. We discard the high-volume ``easy-medium'' mode (grey) to train exclusively on the high-complexity tail (red).}
    \label{fig:deepmath_dist}
\end{figure}

\textbf{Filtering Logic.} We applied a strict threshold of $\ge 8.5$, discarding the vast majority of the dataset to focus exclusively on the ``long tail'' of complexity. As illustrated in Figure \ref{fig:deepmath_dist}, this reduces the dataset from $\sim 103,000$ to 5,399 samples.

\begin{figure}[h]
    \centering
    \includegraphics[width=0.9\linewidth]{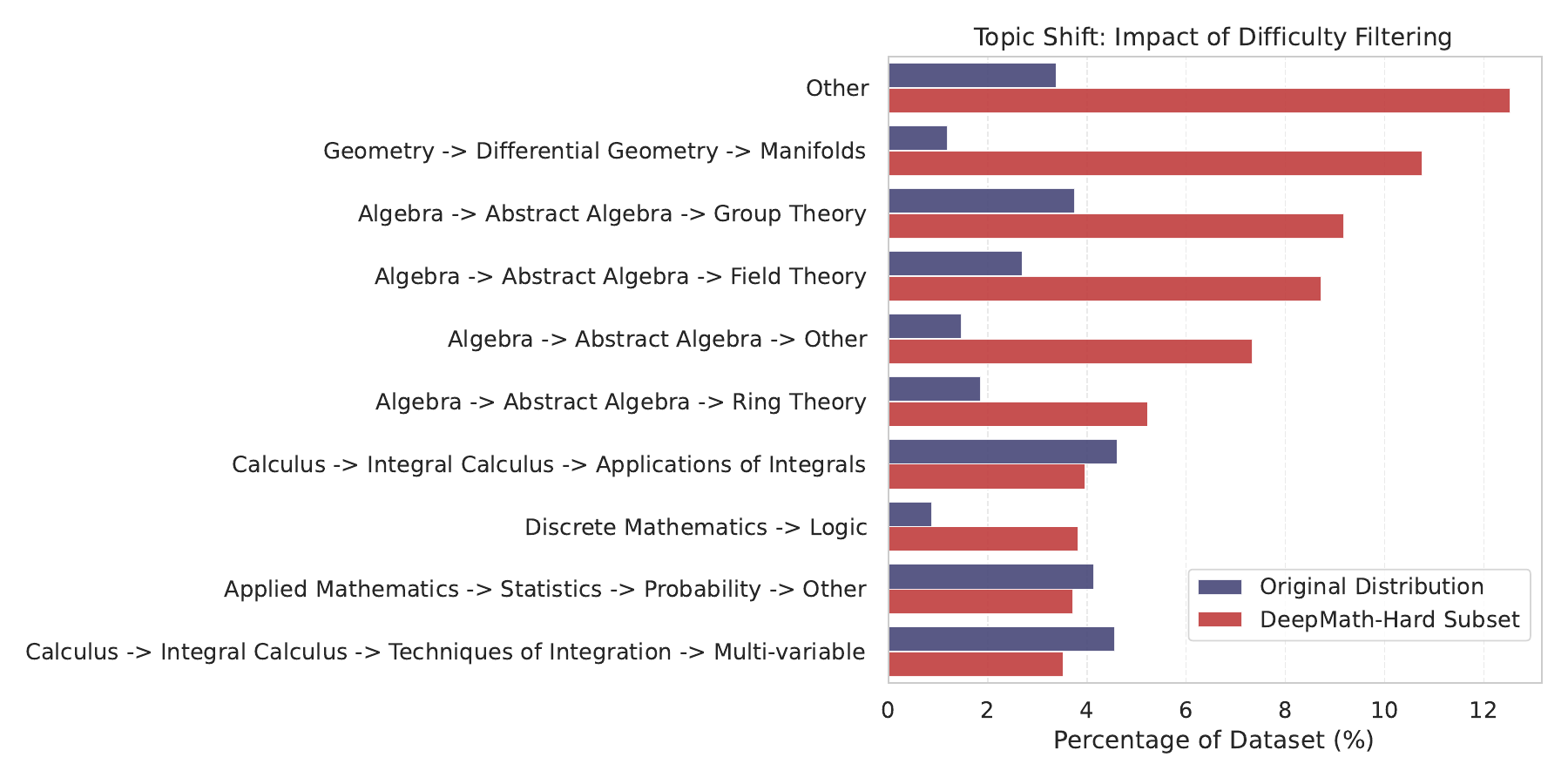}
    \caption{\textbf{Topic Shift Analysis.} Filtering for difficulty ($\ge 8.5$) fundamentally alters the dataset composition, enriching for abstract reasoning domains like Manifolds and Field Theory while suppressing procedural calculus.}
    \label{fig:deepmath_topics}
\end{figure}

\textbf{Impact on Topic Distribution.} Filtering for difficulty implicitly shifts the topic distribution toward abstract mathematical domains. As shown in Figure \ref{fig:deepmath_topics}, the ``Hardcore'' subset (red) is enriched for fields like \textit{Differential Geometry}, \textit{Abstract Algebra (Group/Field Theory)}, and \textit{Topology}, while simpler procedural topics like standard Calculus are de-emphasized compared to the original distribution (purple). This suggests that the filter shifts the training distribution toward problems
requiring more abstract mathematical reasoning.

\subsection{Skywork-Hard and the 32B Judge}
For the \texttt{Skywork-OR1-RL-Data} \cite{Yang2024skywork}, we utilized the \texttt{train-math-deepscaler} subset. Unlike DeepMath, this dataset includes metadata on how larger models performed on each specific problem.

\textbf{The 32B Judge Protocol.} We utilized the \texttt{model\_difficulty} metadata field for the \texttt{DeepSeek-R1-Distill-Qwen-32B} model. This field reports a score inversely proportional to the pass rate. We selected a cutoff score of $\ge 13$, in the Skywork metadata schema, higher scores indicate higher failure rates for the 32B model. 

\begin{figure}[h]
    \centering
    \includegraphics[width=0.8\linewidth]{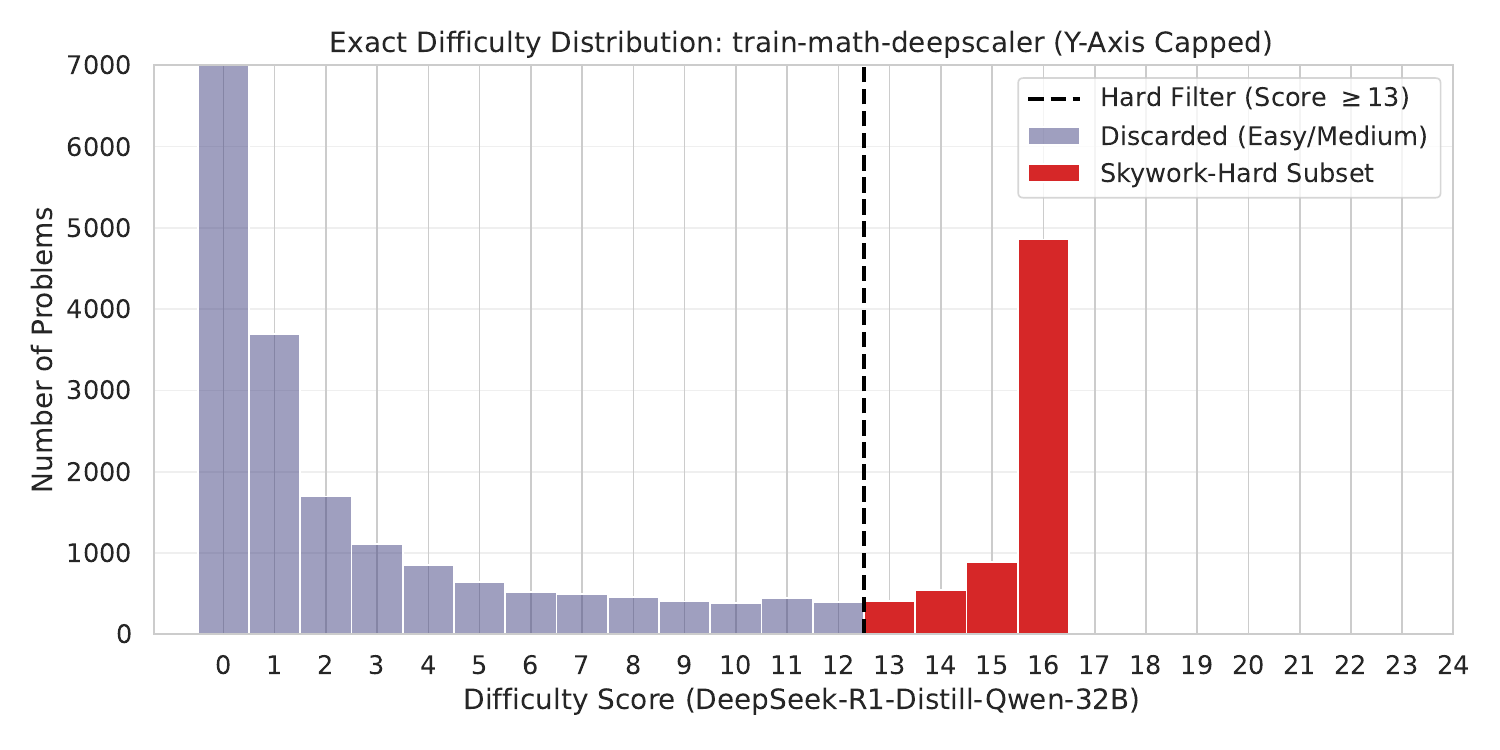}
    \caption{\textbf{The 32B Judge Filter on Skywork-Math.} The y-axis is capped to visualize the tail. The vast majority of problems (Scores 0-12) are discarded. We retain only the subset (red) where the 32B judge model has a low reported pass rate (Score $\ge 13$).}
    \label{fig:skywork_dist}
\end{figure}

\textbf{Distributional Cutoff.} Figure \ref{fig:skywork_dist} visualizes this exact cutoff. The filtering removes the massive volume of ``trivial'' problems and isolates a cluster of 6,702 problems on which the 32B judge model has a low reported pass rate.

\subsection{Enigmata 10k Train and Eval}

We construct a reduced Enigmata training set from the original Enigmata-Data training split released by BytedTsinghua-SIA. The full training split contains 217,541 examples across 38 task names, with a highly non-uniform task distribution. Most tasks contain 6,000 examples each, but several deviate substantially from this mode: \texttt{full\_crosswords} contains 19,000 examples, \texttt{arc\_agi} 11,116, \texttt{pattern\_recognition} 9,000, \texttt{hitori}, \texttt{light\_up}, \texttt{natural\_language\_navigation}, and \texttt{tic\_tac\_toe} 4,000 each, \texttt{knights\_and\_knaves} 3,414, \texttt{checkmate\_in\_one} 2,500, \texttt{big\_bench\_symbolic} 2,000, \texttt{hamiltonian\_path} 1,095, and \texttt{symbolic\_hard} 416. Difficulty labels are similarly imbalanced: among rows with valid difficulty annotations, the original training split contains 56,477 easy, 77,451 medium, and 83,197 hard examples.

\paragraph{Shared-task restriction.}
To ensure that training and evaluation measure performance over the same task family, we first restrict the training pool to tasks shared between the Enigmata training split and the official Enigmata evaluation split. This removes the train-only tasks \texttt{arc\_agi}, \texttt{pattern\_recognition}, and \texttt{symbol\_pattern}. Conversely, the eval-only task \texttt{FOLIO} cannot be included because it is absent from the training split. The resulting candidate pool therefore consists of the 35 task names common to both training and evaluation.

\paragraph{Stratified 10k construction.}
From this shared-task pool, we sample 10,000 examples using task-by-difficulty strata. The sampling procedure enforces two constraints. First, the marginal task distribution of the reduced set matches the relative task frequencies of the original Enigmata training split after restricting to shared tasks. Second, the marginal difficulty distribution matches that of the official evaluation set, rather than the more imbalanced difficulty distribution of the original training split. Concretely, the final Enigmata 10k subset contains 2,930 easy, 3,381 medium, and 3,689 hard examples.

This construction preserves the original train-task mixture as closely as possible while aligning the reduced training set with the evaluation set's difficulty profile. As a result, the Enigmata experiments test generalization over the same task families present at evaluation time without allowing train-only tasks to distort the reduced training distribution. All Enigmata models are trained on this 10k subset and evaluated on the disjoint official Enigmata evaluation set.


\section{Full Rank Sweep Results}
\label{app:rank_sweep}

\subsection{LoRA Rank Sweep with Spectral Diagnostics}

Table~\ref{tab:lora_sweep_full} extends the main-body rank sweep
(Table~\ref{tab:rank_sweep}) with mean effective-rank measurements for every
configuration, enabling direct comparison between nominal and realized update
dimensionality.

\begin{table}[h]
  \centering
  \caption{
    \textbf{Full LoRA rank sweep on Qwen~3~8B, DeepMath-Hard
    (seed 42, batch size 64).}
    Mean effective rank is the participation ratio
    $\rho=\exp(H(\mathbf{p}))$ averaged across target modules.
    Effective rank generally grows with nominal rank in the functional regime,
    but remains small relative to the available rank at high $r$ and does not
    provide a scalar threshold for rank collapse.
  }
  \label{tab:lora_sweep_full}
  \smallskip
  \begin{tabular}{lcccc}
    \toprule
    \textbf{Method} & \textbf{Nominal Rank} & \textbf{Trainable Params}
      & \textbf{Mean Eff. Rank} & \textbf{DeepMath-Hard} \\
    \midrule
    LoRA & $r=1$   & 2.7M   & \phantom{0}1.0000 & 79.29\% \\
    LoRA & $r=4$   & 10.9M  & \phantom{0}2.9191 & 79.29\% \\
    LoRA & $r=8$   & 22M    & \phantom{0}4.709  & 78.1\%  \\
    LoRA & $r=16$  & 43.6M  & \phantom{0}7.1    & 76.2\%  \\
    LoRA & $r=32$  & 87.2M  & 10.2              & 77.3\%  \\
    LoRA & $r=64$  & 174M   & 13.807             & 73.1\%  \\
    LoRA & $r=128$ & 348M   & \phantom{0}9.7    & \phantom{0}4.7\% \\
    LoRA & $r=256$ & 696M   & 17.956             & \phantom{0}2.3\% \\
    \bottomrule
  \end{tabular}
\end{table}

Three aspects of Table~\ref{tab:lora_sweep_full} are worth emphasizing.
First, the low-rank sweep is spectrally coherent: effective rank
increases from 1.000 at $r=1$ to 13.807 at $r=64$, while accuracy remains
broadly stable through $r=32$ before declining at $r=64$.
Thus, increasing effective rank within the functional regime does not
monotonically improve policy quality.

Second, the collapsed high-rank runs rule out a simple effective-rank
threshold for failure.
LoRA $r=128$ collapses with mean effective rank 9.7, while $r=256$ collapses
with effective rank 17.956---higher than the functional $r=64$ checkpoint at
13.807.
Effective rank is therefore informative about the distribution of update
energy, but it does not by itself determine whether optimization succeeds.

Third, realized dimensionality grows far more slowly than nominal capacity.
For example, mean effective rank is approximately 59\% of nominal rank at
$r=8$, but only about 8\% and 7\% at $r=128$ and $r=256$, respectively.
High nominal rank therefore does not translate proportionally into broader
learned updates.
Together with the structured-adapter results in
Section~\ref{sec:experiments:spectra}, this motivates update geometry rather
than nominal or effective rank alone as the more useful organizing principle.

\subsection{DoRA Rank Sweep}

Table~\ref{tab:dora_sweep_full} presents the full DoRA rank sweep on
Qwen~3~8B.
Unlike LoRA, DoRA does not exhibit catastrophic collapse at any rank tested:
performance peaks at $r=16$ (84.0\%) and degrades gracefully to 75.0\% at
$r=64$, with no discontinuous cliff.

\begin{table}[h]
  \centering
  \caption{
    \textbf{DoRA rank sweep on Qwen~3~8B, DeepMath-Hard (seed 42).}
    Graceful degradation at high rank contrasts sharply with LoRA's catastrophic collapse, suggesting that DoRA's
magnitude-direction decomposition acts as a structural regularizer under this RLVR protocol.
  }
  \label{tab:dora_sweep_full}
  \smallskip
  \begin{tabular}{lccc}
    \toprule
    \textbf{Method} & \textbf{Nominal Rank} & \textbf{Trainable Params}
      & \textbf{DeepMath-Hard} \\
    \midrule
    DoRA & $r=8$  & 23.2M & 81.9\% \\
    DoRA & $r=16$ & 45.0M & \textbf{84.0\%} \\
    DoRA & $r=32$ & 90.0M & 80.9\% \\
    DoRA & $r=64$ & 180M  & 75.0\% \\
    \bottomrule
  \end{tabular}
\end{table}

The asymmetry between LoRA and DoRA at high rank is diagnostically
informative.
DoRA's decomposition of each weight matrix into magnitude and direction
components introduces a structural constraint absent from vanilla LoRA.
Under our protocol, this parameterization is associated with graceful
degradation and substantially more distributed learned updates rather than the
catastrophic failure observed for high-rank LoRA.
This is consistent with DoRA's structure regularizing the reachable update
manifold, although the spectral results show that effective rank alone is not
sufficient to explain optimization success.

\subsection{Efficiency Summary}
\label{app:experiments:efficiency}

QuanTA $d=3$'s speed advantage over both $d=4$ and DoRA despite its larger
parameter count reflects the greater parallelism of the $[16,16,16]$ MPO tensor
contraction relative to DoRA's per-column normalization overhead.

\begin{table}[h]
  \centering
  \caption{
    \textbf{Efficiency comparison, Qwen~3~8B on DeepMath-Hard (seed 42).}
    Training time reflects a full single-epoch run on 4$\times$H100.
    In this seed-42 comparison, QuanTA $d=3$ is the strongest efficiency point among the tested configurations:
it matches DoRA $r=16$ on DeepMath-Hard while using fewer trainable parameters and less wall-clock time.
  }
  \label{tab:efficiency}
  \smallskip
  \begin{tabular}{llccc}
    \toprule
    \textbf{Method} & \textbf{Config}
      & \textbf{Trainable Params}
      & \textbf{Training Time}
      & \textbf{DeepMath-Hard} \\
    \midrule
    LoRA   & $r=8$  & 22.0M        & 13.1h           & 78.1\% \\
    LoRA   & $r=16$ & 43.6M        & 12.4h           & 76.2\% \\
    DoRA   & $r=16$ & 45.0M        & 15.6h           & 84.0\% \\
    QuanTA & $d=4$  & 7.7M         & 12.9h           & 78.5\% \\
    QuanTA & $d=3$  & 35.3M        & \phantom{0}7.5h & 84.0\% \\
    \bottomrule
  \end{tabular}
\end{table}


\section{Batch-Size 256 Train-Time Dynamics}
\label{app:bs256_dynamics}

\begin{figure}[h]
  \centering
  \includegraphics[width=0.82\linewidth]{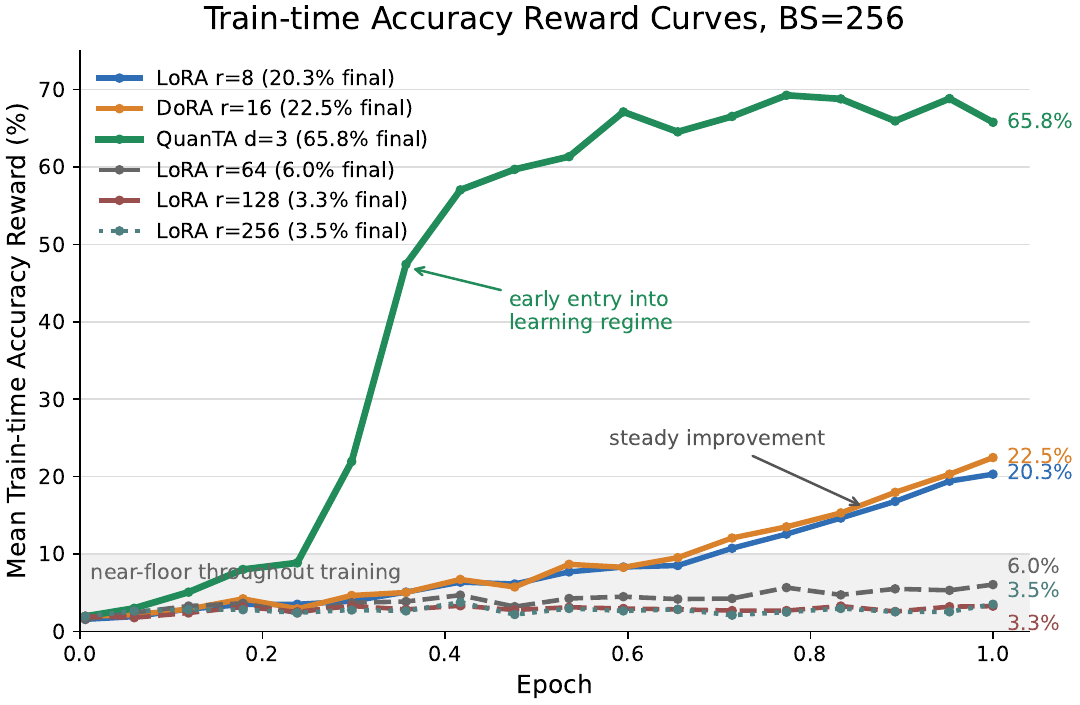}
  \caption{
    \textbf{Train-time verifier accuracy at batch size 256.}
    All curves are Qwen~3~8B on DeepMath-Hard, seed 42, one epoch.
    LoRA $r=8$ and DoRA $r=16$ show steady improvement but remain update-limited
    after 168 optimizer steps.
    QuanTA $d=3$ enters the learning regime early and reaches 65.8\%.
    High-rank LoRA remains near floor throughout training, indicating
    non-ignition rather than slow learning.
  }
  \label{fig:bs256_training_curves}
\end{figure}

Figure~\ref{fig:bs256_training_curves} reports train-time verifier accuracy
curves for the batch-size-256 runs from Section~\ref{sec:experiments:batch_size}.
This setting is intentionally update-limited: under the one-epoch DeepMath-Hard
protocol, batch size 256 yields only 168 optimizer steps, compared with 674
steps at batch size 64.

The curves separate slow learning from non-ignition.
LoRA $r=8$ and DoRA $r=16$ improve steadily throughout training and reach
20.3\% and 22.5\%, respectively, suggesting that these runs are primarily
limited by the reduced number of optimizer updates.
In contrast, LoRA $r=64$, $r=128$, and $r=256$ remain near floor for the entire
run, indicating failure to enter the learning regime rather than merely slower
convergence.
QuanTA $d=3$ exhibits a qualitatively different trajectory: it enters the
learning regime early and reaches 65.8\% despite the same 168-step budget.
This supports the interpretation that structured adapters can widen the stable optimization region in cold-start GRPO, while unconstrained high-rank LoRA can fail to enter the learning regime under the same fixed-pass protocol.

\newpage

\section{Multi-Seed Results}
\label{app:multiseed}

All results in this appendix use seeds 42, 43, and 44.
Standard deviations reflect genuine stochasticity in cold-start GRPO at batch
size 64, not measurement noise.

\subsection{Qwen~3~8B --- DeepMath-Hard}

\begin{table}[h]
  \centering
  \caption{
    \textbf{Multi-seed DeepMath-Hard results, Qwen~3~8B.}
    DoRA, QuanTA, and FFT all exhibit run-to-run variation.
    No method achieves a statistically clean ordering on mean accuracy; the
    primary result is that the structured adapters remain competitive with FFT
    while updating less than 0.6\% of the model parameters.
  }
  \label{tab:deepmath_multiseed}
  \smallskip
  \setlength{\tabcolsep}{4pt}
  \begin{tabular}{llccccc}
    \toprule
    \textbf{Method} & \textbf{Config}
      & \textbf{Seed 42} & \textbf{Seed 43} & \textbf{Seed 44}
      & \textbf{Mean} & \textbf{Std} \\
    \midrule
    FFT    & ---                     & 77.30\% & 77.73\% & 64.45\% & 73.16\% & 7.55\% \\
    \midrule
    DoRA   & $r=16$                  & 84.0\%  & 76.6\%  & 62.5\%  & 74.4\%  & 10.9\% \\
    DoRA   & $r=16$ (\texttt{no\_kv}) & 80.9\%  & 77.0\%  & 60.9\%  & 72.9\%  & 10.4\% \\
    \midrule
    QuanTA & $d=3$ (\texttt{linear})      & 78.5\% & 79.3\% & 64.5\% & 74.1\% & 8.4\% \\
    QuanTA & $d=3$ (\texttt{no\_v\_proj}) & 80.1\% & 78.5\% & 63.7\% & 74.1\% & 9.0\% \\
    QuanTA & $d=3$ (\texttt{no\_kv})      & 84.0\% & 76.2\% & 65.2\% & 75.1\% & 9.6\% \\
    \bottomrule
  \end{tabular}
\end{table}

The three-seed comparison places the non-collapsed methods in the same broad
performance range. QuanTA $d=3$ (\texttt{no\_kv}) has the highest observed
mean at 75.1\%, followed by DoRA $r=16$ at 74.4\% and FFT at 73.16\%, but
these differences are small relative to the observed run-to-run variation.
We therefore do not claim a statistically clean performance ordering among
DoRA, QuanTA, and FFT. The main result is instead that the structured adapters
remain competitive with full fine-tuning while using orders of magnitude fewer
trainable parameters.

Seed 44 is the lowest-performing seed for FFT and for nearly all structured
configurations shown here, suggesting that part of the observed variance is
shared across methods rather than being specific to one parameterization.

\subsection{Qwen~3~8B — Skywork-Hard}

\begin{table}[h]
  \centering
  \caption{
    \textbf{Multi-seed Skywork-Hard results, Qwen~3~8B.}
    High problem difficulty produces substantial reward variance throughout
    training, which propagates into evaluation variance.
    Differences between methods are within noise; the primary signal is that
    all methods remain functional on this benchmark, unlike the catastrophic
    collapses observed at high LoRA rank.
  }
  \label{tab:skywork_multiseed}
  \smallskip
  \begin{tabular}{llccccc}
    \toprule
    \textbf{Method} & \textbf{Config}
      & \textbf{Seed 42} & \textbf{Seed 43} & \textbf{Seed 44}
      & \textbf{Mean} & \textbf{Std} \\
    \midrule
    LoRA   & $r=8$ & 19.0\% & 13.2\% & 21.4\% & 17.9\% & 4.3\% \\
    QuanTA & $d=4$ & 19.6\% & 11.2\% & 18.5\% & 16.4\% & 4.6\% \\
    QuanTA & $d=3$ & 19.4\% & 14.5\% & 21.7\% & 18.5\% & 3.6\% \\
    \bottomrule
  \end{tabular}
\end{table}

QuanTA $d=3$ achieves the highest mean (18.5\%) and the lowest standard deviation (3.6\%) among the
tested configurations, but the differences are small relative to the observed variance. We therefore interpret
Skywork-Hard mainly as evidence that all tested non-collapsed configurations remain functional on this
harder benchmark.

\subsection{Llama~3.1~8B — MATH}

Table \ref{tab:llama_multiseed} shows a sharp seed-level instability for QuanTA $d=4$ on Llama 3.1 8B: seed 43 collapses to
0.8\%, while LoRA $r=16$ remains stable across all three seeds. We do not discard this run as an outlier,
because such instability is informative under the Model Maturity Hypothesis. One interpretation is that,
for less mature models, structured high-rank adaptation may expose optimization directions that are not
reliably useful under sparse binary rewards, whereas LoRA's lower-rank constraint provides stronger
regularization. This remains an interpretation rather than a causal conclusion; more models and seeds would
be required to determine whether this is a general property of less mature base models.
\begin{table}[h]
  \centering
  \caption{
    \textbf{Multi-seed MATH results, Llama~3.1~8B.}
    LoRA $r=16$ is stable across all seeds; QuanTA $d=4$ collapses entirely on
    seed 43 (0.8\%), producing high variance and a substantially lower mean.
    This asymmetry---stable under LoRA, volatile under structured high-rank adaptation---is one behavioral
signature motivating the Model Maturity Hypothesis for less mature models.
  }
  \label{tab:llama_multiseed}
  \smallskip
  \begin{tabular}{llccccc}
    \toprule
    \textbf{Method} & \textbf{Config}
      & \textbf{Seed 42} & \textbf{Seed 43} & \textbf{Seed 44}
      & \textbf{Mean} & \textbf{Std} \\
    \midrule
    LoRA   & $r=16$ & 28.9\% & 34.3\% & 27.3\% & 30.2\% & \phantom{0}3.7\% \\
    QuanTA & $d=4$  & 26.6\% & \phantom{0}0.8\% & 21.9\% & 16.4\% & 13.7\% \\
    \bottomrule
  \end{tabular}
\end{table}

\section{Constant Learning Rate Collapse}
\label{app:lr_collapse}

\begin{figure}[h]
    \centering
    \includegraphics[width=0.5\linewidth]{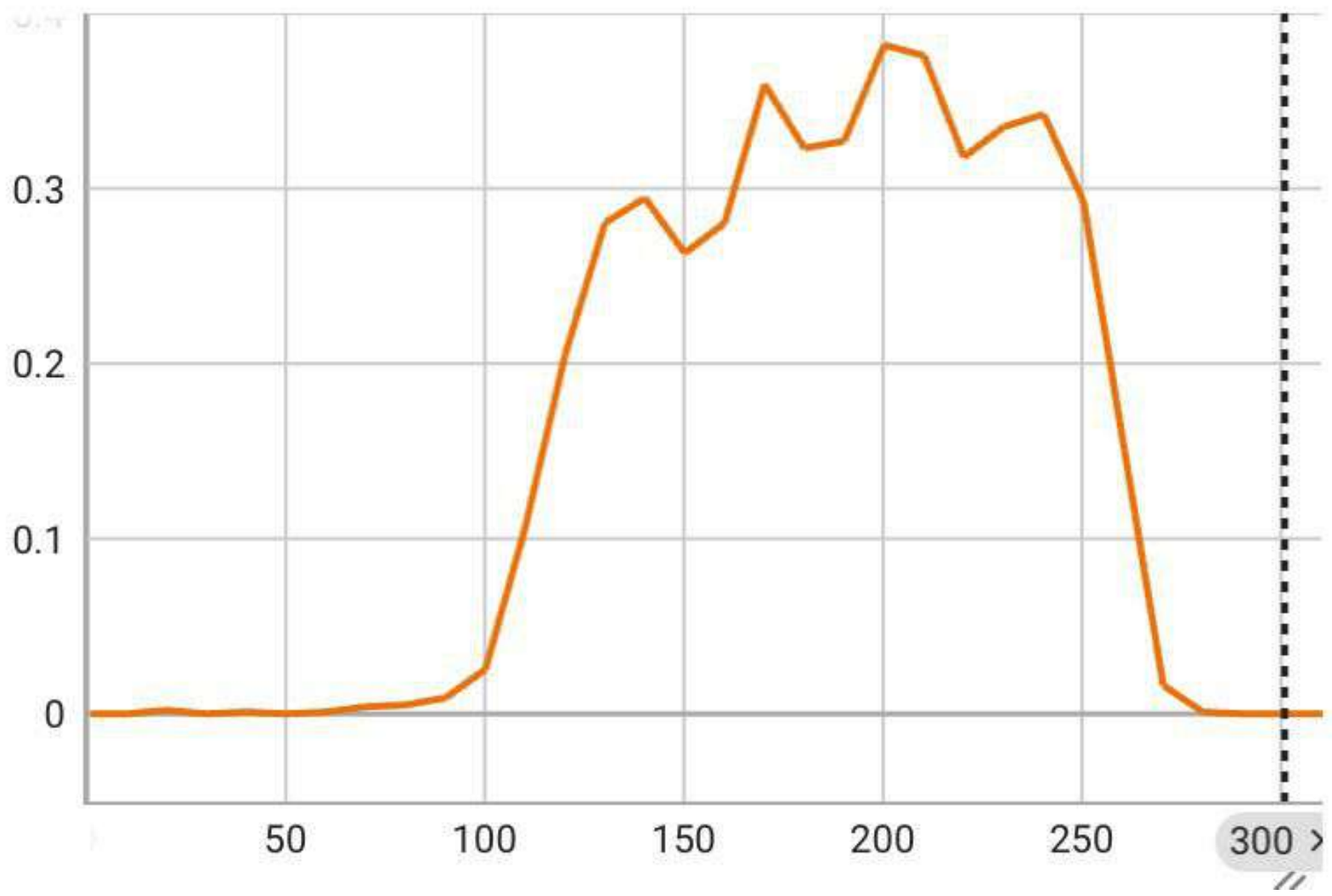}
    \caption{Training reward under a constant learning rate schedule (QuanTA $d=3$, 
    Qwen-3 8B, MATH dataset). The policy collapses catastrophically near step 280 despite healthy 
    early progress, illustrating why cosine decay with warmup is suggested 
    in cold-start GRPO at batch size 64. The dashed vertical line marks the early stoppage of 
    training.}
    \label{fig:lr_collapse}
\end{figure}

Figure \ref{fig:lr_collapse} shows the training reward curve of QuanTA $d=3$ on Qwen-3 8B trained on MATH under a
constant learning-rate schedule, with all other hyperparameters identical to the main experiments. The
policy initially learns, peaking near step 200, before collapsing to near-zero reward by step 280. We interpret
this failure as arising from the interaction between sparse rewards, batch size 64, and the absence of
learning-rate decay. Without decay, the optimizer continues to take large updates late in training, when
advantage estimates can become unstable, and the policy fails to recover. A cosine schedule with warmup
ratio 0.1 and minimum learning-rate ratio 0.15 avoids this failure mode in our main protocol, so we adopt it
as a fixed component rather than a tuned hyperparameter.

\section{QuanTA Target-Module Ablation}
\label{app:module_ablation}

Table~\ref{tab:module_ablation} reports DeepMath-Hard results for QuanTA across
three target-module configurations: \texttt{target\_linear} (all linear
modules), \texttt{no\_v\_proj} (all linear modules except the value
projection), and \texttt{no\_kv} (all linear modules except key and value
projections).

\begin{table}[h]
  \centering
  \caption{
    \textbf{QuanTA target-module ablation on Qwen~3~8B, DeepMath-Hard.}
    Excluding key and value projections consistently produces the best or
    near-best single-seed performance for $d=3$, and the best mean across seeds.
    $d=4$ multi-seed results are not available; seed-42 values are reported
    for reference.
  }
  \label{tab:module_ablation}
  \smallskip
  \setlength{\tabcolsep}{4pt}
  \begin{tabular}{llcccccc}
    \toprule
    \textbf{Config} & \textbf{Modules}
      & \textbf{Seed 42} & \textbf{Seed 43} & \textbf{Seed 44}
      & \textbf{Mean} & \textbf{Std} \\
    \midrule
    $d=3$ & target\_linear & 78.5\% & 79.3\% & 64.5\% & 74.1\% & 8.4\% \\
    $d=3$ & no\_v\_proj    & 80.1\% & 78.5\% & 63.7\% & 74.1\% & 9.0\% \\
    $d=3$ & no\_kv         & \textbf{84.0\%} & 76.2\% & 65.2\% & \textbf{75.1\%} & 9.6\% \\
    \midrule
    $d=4$ & target\_linear & 79.7\% & --- & --- & --- & --- \\
    $d=4$ & no\_kv         & 78.5\% & --- & --- & --- & --- \\
    \bottomrule
  \end{tabular}
\end{table}

The no\_kv configuration for $d=3$ achieves both the best seed-42 result and the best mean across seeds,
although its mean advantage over the other $d=3$ variants is small relative to the standard deviation. This
suggests that excluding key and value projections does not harm performance in this setting and may provide
a mild regularizing effect. One possible explanation is that key and value projections encode content-retrieval
representations that are already useful from pre-training, so adapting them aggressively during reasoning
alignment can be unnecessary or mildly harmful. Under this interpretation, query projections and feed-forward
layers provide a more useful locus for reasoning-oriented adaptation in the no\_kv configuration. For $d=4$,
the target\_linear configuration marginally outperforms no\_kv on seed 42 (79.7\% vs. 78.5\%), but the
difference is small and multi-seed validation is unavailable; we therefore avoid making a definitive target-module
recommendation for $d=4$.

\section{AIME 2025--2026 Generalization}
\label{app:aime}

\begin{table}[h]
  \centering
  \caption{
    \textbf{AIME 2025--2026 out-of-distribution generalization (Avg@32),
    Qwen~3~8B.}
    All models trained on DeepMath-Hard only (${\sim}5{,}400$ samples).
    Full fine-tuning collapses to near-random performance despite achieving
    competitive in-distribution accuracy on DeepMath-Hard (77.3\%,
    Table~\ref{tab:main_results}).
  }
  \label{tab:aime}
  \smallskip
  \begin{tabular}{llc}
    \toprule
    \textbf{Method} & \textbf{Config} & \textbf{AIME Avg@32} \\
    \midrule
    LoRA   & $r=16$         & \phantom{0}9.95\%  \\
    DoRA   & $r=16$         & \textbf{11.04\%} \\
    QuanTA & $d=3$ (no\_kv) & 10.36\% \\
    \midrule
    FFT    & ---            & \phantom{0}1.35\% \\
    \bottomrule
  \end{tabular}
\end{table}

FFT's near-floor AIME performance (1.35\%) is notable because the same
checkpoint remains competitive in-distribution on DeepMath-Hard
(77.3\%, Table~\ref{tab:main_results}), close to LoRA $r=8$ at 78.1\%.
This makes a simple in-distribution memorization explanation insufficient:
FFT does not substantially outperform LoRA on DeepMath-Hard, yet transfers far
more poorly to the AIME distribution. We therefore interpret the result as
suggestive evidence that the learned FFT update may be less robust
out-of-distribution despite fitting the DeepMath-Hard regime comparably well.
However, the merged AIME 2025--2026 evaluation contains only 60 problems, so
this observation should be treated as suggestive rather than definitive.

A possible explanation is the effective-rank pattern documented in Section 4.5. FFT updates all parameters
but produces lower effective-rank updates than the structured adapters. Without an explicit structural prior,
these updates may become more tied to DeepMath-Hard-specific features, such as topic distribution, phrasing,
or solution style, and less robust to AIME's shorter competition-style problems. LoRA, DoRA, and QuanTA
all score in a narrow 10--11\% range, so the main AIME signal is not a clean ordering among adapters; rather,
it is that FFT generalizes unusually poorly in this OOD probe. This is consistent with the broader claim that
unconstrained optimization under sparse rewards can produce brittle updates, but additional OOD benchmarks
and seeds would be needed to establish the mechanism.


\section{Frozen Weight Spectral Analysis}
\label{app:frozen_spectra}

\begin{figure}[h!]
  \centering
  \includegraphics[width=0.85\linewidth]{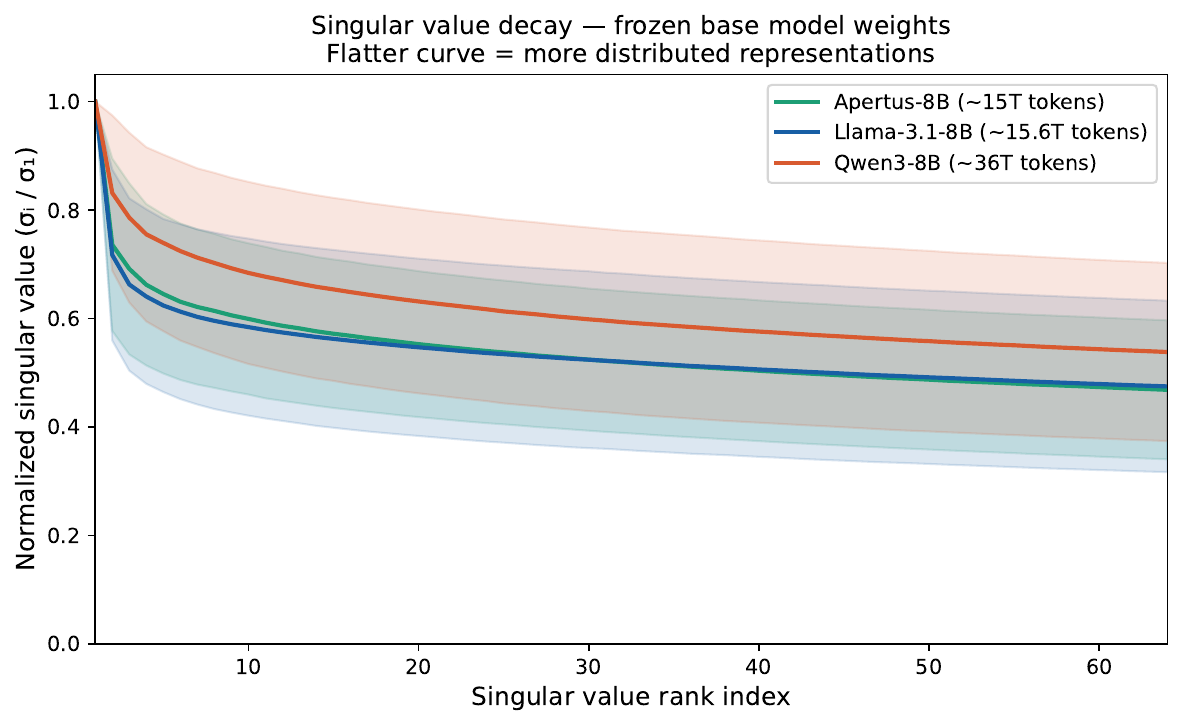}
  \caption{
    \textbf{Singular value decay of frozen base model weights, averaged across
    sampled linear layers using the top 64 singular values per layer.} Shaded regions denote one standard deviation across
layers. Qwen 3 8B ($\sim$36T tokens) exhibits markedly flatter decay than both $\sim$15T models. Apertus-8B
and Llama 3.1-8B produce closely overlapping decay curves despite differing model families. This frozen-weight
spectral signature is consistent with the behavioral partition observed in Section \ref{sec:experiments:maturity}, but does not by itself
causally isolate pre-training scale from other model differences.
  }
  \label{fig:frozen_decay}
\end{figure}

Figure \ref{fig:frozen_decay} plots the normalized singular value decay of frozen linear layers across all three base models,
measured before any fine-tuning occurs. This analysis provides a training-free spectral correlate of the
behavioral partition documented in Section \ref{sec:experiments:maturity}.

Three observations follow from Figure \ref{fig:frozen_decay}.

\paragraph{The maturity ordering is visible in frozen weights.}
Qwen 3 8B's decay curve sits consistently above the two $\sim$15T models across all 64 singular-value
indices, indicating flatter spectra in its pre-trained weights. Because this measurement is taken before
fine-tuning, it provides a candidate pre-training diagnostic associated with the later adaptation behavior.

\paragraph{Apertus and Llama are spectrally similar.}
Despite differing in architecture, tokenizer, and training data mixture, the two $\sim$15T models produce
overlapping decay curves with nearly identical means throughout. This reduces the likelihood that the
observed behavioral partition is purely family-specific, but it does not rule out all confounds: architecture,
tokenizer, data mixture, and pre-training data volume still co-vary in the current model set.

\paragraph{Proposed mechanism.}
We hypothesize that steep singular value decay corresponds to more concentrated pre-trained representations,
for which low-rank updates can provide useful regularization. Conversely, flatter spectra may indicate that
useful representations are distributed across more directions, making adaptation benefit from updates spread
across a broader singular subspace. Under this interpretation, Qwen 3 8B benefits more clearly from structured
high-rank adaptation because QuanTA can express distributed updates while still constraining the reachable
update manifold. This mechanism is consistent with the observed performance gap on Qwen 3 8B and the
absence of a comparable gap on the two $\sim$15T models, but controlled experiments with matched
architectures and intermediate pre-training checkpoints are needed to establish causality.

\section{Qualitative Analysis: An Illustrative Reasoning Failure}
\label{app:qualitative}

To illustrate the kinds of reasoning differences that can accompany the quantitative gaps, we analyze one
Chain-of-Thought trace generated by Qwen 3 8B on a complex counting problem from Skywork-Hard. This
case study is not intended as standalone evidence for the mechanism; rather, it provides a concrete example
of the failure modes that motivate further qualitative analysis.

We compare a LoRA $r=16$ checkpoint against the QuanTA $d=3$ checkpoint sampled from the final 5\%
of the training run.

\subsection{Case Study: Combinatorial Counting}

\textbf{Problem:} \textit{In an Olympic 100-meter final, there are 10 sprinters competing, among which 4 are Americans. The gold, silver, and bronze medals are awarded to first, second, and third place, respectively. Calculate the number of ways the medals can be awarded if at most two Americans are to receive medals.}

\textbf{Correct Answer:} 588.

\begin{table}[h!]
\centering
\begin{tabular}{p{0.48\textwidth}|p{0.48\textwidth}}
\toprule
\multicolumn{1}{c|}{\textbf{LoRA ($r=16$) Trace [FAILURE]}} & \multicolumn{1}{c}{\textbf{QuanTA ($d=3$) Trace [SUCCESS]}} \\
\midrule
\textit{... (Case 1: No Americans is correct) ...} & \textit{... (Case 1: No Americans is correct) ...} \\
\addlinespace[0.5em]
\textbf{Case 2: Exactly one American.} & \textbf{Case 2: Exactly one American.} \\
``We have 4 choices for which American receives a medal. For that American, we have 4 choices (since they can get any of the 4 positions). \textcolor{red}{[ERROR: There are only 3 medal positions, not 4]} ... The number of ways is $4 \times 4 \times 6 \times 5 = 480$.'' & ``In this case, we have to choose which of the three medals the American will receive, then choose which one of the 4 Americans will receive that medal... The number of ways is: $3 \times 4 \times P(6, 2) = 3 \times 4 \times 30 = \mathbf{360}$.'' \\
\addlinespace[0.5em]
\textbf{Case 3: Exactly two Americans.} & \textbf{Case 3: Exactly two Americans.} \\
``...For the remaining medal, we have 6 choices. The number of ways to award the medals in this case is: $6 \times 6 \times 6 = 216$.'' \textcolor{red}{[ERROR: Failed to account for permutations of the specific Americans selected]} & ``...Choose which two of the three medals the Americans will receive, then choose which two of the 4 Americans... The number of ways is: $3 \times \binom{4}{2} \times 6 = 3 \times 6 \times 6 = \mathbf{108}$.'' \\
\addlinespace[0.5em]
\textbf{Final Calculation:} & \textbf{Final Calculation:} \\
$120 + 480 + 216 = \mathbf{816}$ & $120 + 360 + 108 = \mathbf{588}$ \\
\bottomrule
\end{tabular}
\caption{\textbf{Illustrative reasoning failure.} LoRA (left) identifies the correct case structure but makes local
counting errors in Case 2 and Case 3. QuanTA (right) correctly applies the relevant permutation and binomial
factors for this example. This single trace illustrates a possible state-tracking failure mode, but should not be
read as a statistically representative qualitative evaluation.}
\label{tab:qualitative_comparison}
\end{table}

\subsection{Analysis of the Failure}

The LoRA trace retrieves the high-level strategy---case analysis by number of American medalists---but
fails in the local combinatorial accounting. In Case 2, it refers to ``4 positions'' even though there are only
three medals. In Case 3, it defaults to a simple multiplication heuristic and fails to account for the dependency
between selected Americans and medal slots. These errors are consistent with a loss of state precision during
multi-step reasoning.

The QuanTA trace preserves the relevant state variables in this example: which medals are assigned, which
Americans are selected, and how many non-American sprinters remain. This is qualitatively consistent with
the hypothesis that structured high-rank adaptation can better support distributed state tracking. However,
because this appendix presents a single example, we treat it as illustrative rather than conclusive evidence.
A larger qualitative analysis would be needed to determine whether this failure mode is systematically more
common for LoRA.

\section{Conceptual Geometric Comparison of PEFT Methods}
\label{app:peft_theory}

Our spectral analysis suggests two properties that may be useful for strong performance in mature-model
RLVR: (a) high effective rank in the learned $\Delta W$, and (b) structural restriction of the update manifold
so that unconstrained singular-direction concentration is less likely. We analyze several PEFT methods
through this lens. This appendix is conceptual and hypothesis-generating: except for LoRA, DoRA, QuanTA, and FFT, the methods discussed here are not empirically evaluated in our RLVR pipeline.

\paragraph{LoRA.}
Constraining $\Delta W = BA$ to rank at most $r$ bounds the maximum rank, but it does not force the
optimizer to use all available directions. In our RLVR experiments, high-rank LoRA checkpoints
($r \geq 128$) concentrate update energy into a small number of singular directions despite large nominal
rank. LoRA therefore satisfies neither property in this framework: low-rank LoRA lacks high effective rank,
while high-rank LoRA lacks a structural constraint that prevents singular-direction concentration.

\paragraph{DoRA.}
DoRA decomposes each weight into magnitude and direction components, with LoRA applied to the
directional component. In our experiments, this parameterization is associated with substantially higher
effective rank than vanilla LoRA at comparable nominal rank and avoids the catastrophic collapse observed
in high-rank LoRA. We interpret this as evidence that the magnitude-direction decomposition acts as a
regularizer on the reachable update manifold, though the precise causal pathway remains to be established.

\paragraph{QuanTA.}
Parameterizing $\Delta W$ as an MPO tensor network constrains updates to a tensor-structured manifold
$\mathcal{T}_\chi$ whose elements can be high-rank as matrices. Unlike the LoRA rank-$r$ manifold
$\mathcal{M}_r$, this parameterization can distribute update energy across many mode indices while still
restricting the set of reachable matrices. In our experiments, QuanTA $d=3$ achieves the highest measured
effective rank among the tested methods and matches DoRA on the primary DeepMath-Hard seed-42 result.
This supports the view that structured high-rank adaptation, rather than tensor decomposition specifically,
is the important property.

\paragraph{BOFT.}
BOFT restricts updates through butterfly-factored orthogonal transformations. Because orthogonal updates
are norm-preserving, this structure may regularize some forms of unstable rescaling. However, the same
constraint may also limit the ability to change feature magnitudes, which our spectral interpretation suggests
could matter for mature-model RLVR. Since we do not evaluate BOFT empirically, we treat this as a
geometric prediction rather than a conclusion.

\paragraph{MoRA.}
MoRA sandwiches a learnable square matrix between fixed non-learnable compression operators, constraining
updates to a subspace determined before training. This can in principle produce higher-rank updates than
standard LoRA, but the fixed compression structure may be less adaptive to the particular directions needed
during cold-start GRPO. Because MoRA is not evaluated in our experiments, we present this as a hypothesis
about its geometry rather than a claim about its empirical RLVR performance.

\paragraph{Summary of the proposed geometric lens.}
\begin{center}
\begin{tabular}{lcc}
  \toprule
  Method & (a) High effective rank expected & (b) Structural restriction expected\\
  \midrule
  LoRA  & \texttimes & \texttimes \\
  DoRA  & \checkmark & \checkmark \\
  QuanTA & \checkmark & \checkmark \\
  BOFT  & \texttimes & \checkmark \\
  MoRA  & $\sim$ & \texttimes \\
  FFT   & \texttimes & \texttimes \\
  \bottomrule
\end{tabular}
\end{center}

The resulting prediction is testable: PEFT methods that combine high effective rank with a structured
reachable-update manifold should be more stable than unconstrained high-rank LoRA in mature-model
cold-start GRPO. DoRA and QuanTA provide two empirical examples consistent with this prediction, but
additional methods such as BOFT and MoRA would need to be evaluated directly before assigning them a
definitive performance tier.


\newpage

\end{document}